\documentclass[9pt,a4paper,twocolumn, DIV13]{scrartcl}

\usepackage{microtype}
\usepackage{graphicx}
\usepackage[dvipsnames]{xcolor}
\usepackage{subfigure}
\usepackage{booktabs} % for professional tables

\usepackage{hyperref}
\usepackage{url}
\usepackage{booktabs}
\usepackage{multirow}
\usepackage{graphicx}
\usepackage{amsmath,amssymb,mathtools,bm,bbm}
\usepackage{enumerate}
\usepackage{pgfplots}
\usepackage{hyperref}
\usepackage{authblk}
\usepackage[superscript]{cite}
\usepackage{icomma}
\usepackage[font=footnotesize,
	    labelfont=bf,
	    labelfont+=sf,
	    format=plain,
	    indention=4ex]%
	    {caption}

\newcommand{\eg}{e.\,g.}
\newcommand{\FIG}{Fig.}
\newcommand{\FIGS}{Figs.}
\newcommand{\SEC}{Sec.}

\newcommand{\EQ}{Eq.}
\newcommand{\etal}{\textit{et~al.}}

\newcommand{\domA}{\mathcal{A}}
\newcommand{\domB}{\mathcal{B}}

\begin{document}

\title{Unpaired High-Resolution and Scalable Style Transfer \\
Using  Generative Adversarial Networks}
\author{Andrej Junginger}
\author{Markus Hanselmann}
\author{Thilo Strauss}
\author{Sebastian Boblest}
\author{Jens Buchner}
\author{Holger Ulmer}
\affil{Machine Learning Team at ETAS GmbH (Bosch Group), Stuttgart, Germany}
\date{}

\twocolumn[
\begin{@twocolumnfalse}

\maketitle
\begin{abstract}
Neural networks have proven their 
capabilities by outperforming many other approaches on regression or 
classification tasks on various kinds of data.
Other astonishing results have been achieved using neural nets as data
generators, especially in settings of generative adversarial networks (GANs).
One special application is the field of image domain translations. 
Here, the goal is to take an image with a certain style (\eg~a photography) and 
transform it into another one (\eg~a painting).
If such a task is performed for \emph{unpaired} training examples, the 
corresponding GAN setting is complex, the neural networks are large, and 
this leads to a high peak memory consumption during, both, training and 
evaluation phase.
This sets a limit to the highest processable image size.
We address this issue by the idea of not processing the whole image at once, 
but to train and evaluate the domain translation on the level of overlapping 
image subsamples.
This new approach not only enables us to translate high-resolution images that 
otherwise cannot be processed by the neural network at once, but also allows us 
to work with comparably small neural networks and with limited hardware 
resources.
Additionally, the number of images required for the training process is 
significantly reduced.
We present high-quality results on images with a total resolution of up to over 
50 megapixels and demonstrate that our method helps to preserve local image 
details while it also keeps global consistency.
\end{abstract}
\vspace{2em}
\end{@twocolumnfalse}
]

\section{Introduction}

Over the recent years, neural networks (NNs) have become state of the art in 
processing various kinds of data types, such as, \eg, high-dimensional 
numerical 
data \cite{}, images \cite{}, time series \cite{}, or language data 
\cite{jones1995,collobert2011,martinez2013,cambria2012,lawrence2000}, 
only to name a few.
Standard applications are classification or regression tasks, and in many 
cases NNs are oußtperforming classical approaches significantly.
Also in the field of anomaly detection 
\cite{taylor2016,sakurada2014,zimek2012,erfani2016} 
and object recognition 
\cite{felzenszwalb2010,viola2001,sung2002,ohnbar2016,wojek2012,%
kobatake1996,bai2010}, 
NNs have been proven to be powerful approaches.

Recently, it has been shown that NNs can be used also as data generators, 
especially in settings of generative adversarial networks (GANs) 
\cite{goodfellow2014,denton2015,radford2015,salimans2016,zhao2016}.
Therein, two networks -- the generator and the discriminator -- compete with 
each other in a way that the generator learns to generate synthetic data that 
exhibits the specific properties and characteristics of the training data.
% One such data sample can finally be constructed out of a random number which 
% follows a predefined distribution.
A similar task can also be performed by variational autoencoders 
\cite{kingma2014}.

Although the mathematical background of NNs has been known for decades, some of 
the biggest development steps and successful applications have been presented 
only in the recent years.
These breakthroughs are mainly due to two reasons:
On the one hand, we are today equipped with the required computational power, 
especially in the form of GPUs, in order to perform the network training on a 
reasonable time scale.
On the other hand, substantial knowledge about network architectures has been 
developed, \eg~concerning convolutional (CNN) or recurrent (RNN) neural 
networks.

A special field of application which could not be thought of without the 
progress mentioned above is that of style transfer networks
\cite{gatys2016,pix2pix,CycleGAN,UNIT,MUNIT,huang2017,sanakoyeu2018,wang2017,
li2018,ledig2016}.
These algorithms have been developed with special regard to the data type of 
digital images.
Their task is to translate an image of a certain domain 
$\domA$ (\eg~a photo) into the style of a different domain 
$\domB$ (\eg~an artistic painting).
This problem can be faced from different perspectives and in the literature, 
one can find methods using direct optimization procedures 
\cite{gatys2016}, %Tü
methods working on paired images
\cite{pix2pix}, % pix2pix?
and approaches performing unpaired image translations
\cite{CycleGAN,UNIT,MUNIT} % CycleGAN, UNIT
(see also \SEC~\ref{sec:related-work}).
Unpaired frameworks have the advantage that they do not require one-to-one 
training examples from both domains, which are often not easily available or do 
not even exist.
Instead, they use a special kind of NN arrangement in an extended GAN setting 
which makes it possible to train the translation using unpaired image examples.

Unpaired domain translation settings are also subject of this paper and we focus 
on the special case of high-resolution images by which we understand the 
several to high megapixel regime.
This is in contrast to many previous machine learning papers on image tasks 
which have demonstrated applications on publicly available data sets.
Typical representatives of these image data sets are 
MNIST ($28 \times 28$ pixels), 
CIFAR10 ($32 \times 32$ pixels), 
CALTEC101 (about $300 \times 200$ pixels), or others with typical image 
resolutions on the order of some hundred to some ten thousands of pixels 
altogether.
% Few papers treat also slightly higher image resolutions \cite{}.

However, these data sets do by far not reach resolutions that are typical for 
today's camera systems.
Even simple smartphone cameras easily reach the double-digit megapixel 
regime and they can record videos in Full HD (about 2 megapixels).
Video systems with 4K resolution (about 8 megapixels) are commercially 
available and today's standard of DSLR cameras is on the order of 20 megapixels 
and above.
Today, there is a clear demand for such highly resolved images to capture 
relevant image details.
As an example, video systems developed for the use case of autonomous driving 
work on high-resolution images to provide a sufficiently detailed view of the 
car's surrounding.
These numbers demonstrate the clear need to develop machine learning algorithms 
that are capable to handle today's high-resolution image data.

In this paper, we address the problem of unpaired domain translation 
with special 
regard to this issue of being able to process high-resolution images.
We discuss that current methods suffer from a high peak memory 
consumption during training and translation which sets a natural limit to the 
largest processable image size on a given GPU.
To solve this issue, we introduce a scalable method which is able to 
work on arbitrary-high resolutions without increasing the 
peak memory consumption of the NN.
We achieve this goal by the simple idea, not to process the whole image at 
once but to train and apply the domain translation on the level of small, 
overlapping image subsamples.
For the training of the underlying generators, each of the existing methods 
can be applied, and we use a \textsc{Unit}-like 
framework\cite{UNIT} for our investigations in this paper.

A question arising with the task of image translation is how the styles 
of the domains $\domA$ and $\domB$ are defined.
Since there is a high variability in real-world data samples, NNs are usually
trained on huge data sets representing this high variance.
% Usually, NNs are trained in a way that they are able to cover a high variance 
% of possible real-world input data samples.
% This demand is often tried to satisfy using large training data sets with huge 
% variants of examples.
% 
By contrast, there can also be the need to handle the opposite case of low 
variance data:
An example is the case that one of the domains corresponds to simulated images.
Such simulated images often work on textures with low spacial variability. 
As a consequence, different images in the domain change their content but only 
hardly vary in their appearance.
This question is closely related to our goal of developing a high-resolution 
style transfer, in the sense that one large image with all its details may 
contain a variance that is similar to the one of a big data set of small images.
This reflects the fact that the actual relevant measure of the data set used 
for training is less the size in terms of megapixel but rather in the sense of 
its Shannon entropy.

Our paper is organized as follows:
In \SEC~\ref{sec:related-work}, we provide a brief overview of some selected 
style transfer algorithms and discuss some of their advantages and 
shortcomings.
In \SEC~\ref{sec:method}, we introduce our method and explain how we work on 
the level of image subsamples.
Results on high-resolution images are presented in \SEC~\ref{sec:results} with 
examples covering the range from similar to different domains.
We demonstrate that our method works well even for ``single-shot'' 
translations, where the target style is defined by only \emph{one} image, 
and we present results obtained from images with up to more than 50 megapixels 
resolution.
After concluding in \SEC~\ref{sec:conclusion}, we provide additional 
information like the NN's architectural design and image details in the 
appendix.

\section{Related Work}
\label{sec:related-work}

%%%%%%%%%%%%%%%%%%%%%%%%%%%%%%%%%%%%%%%%%%%%%%%%%%%%%%%%%%%%%%%%%%%%%%%%%%%%%%%%
\begin{figure}[t]
\includegraphics[width=\columnwidth]{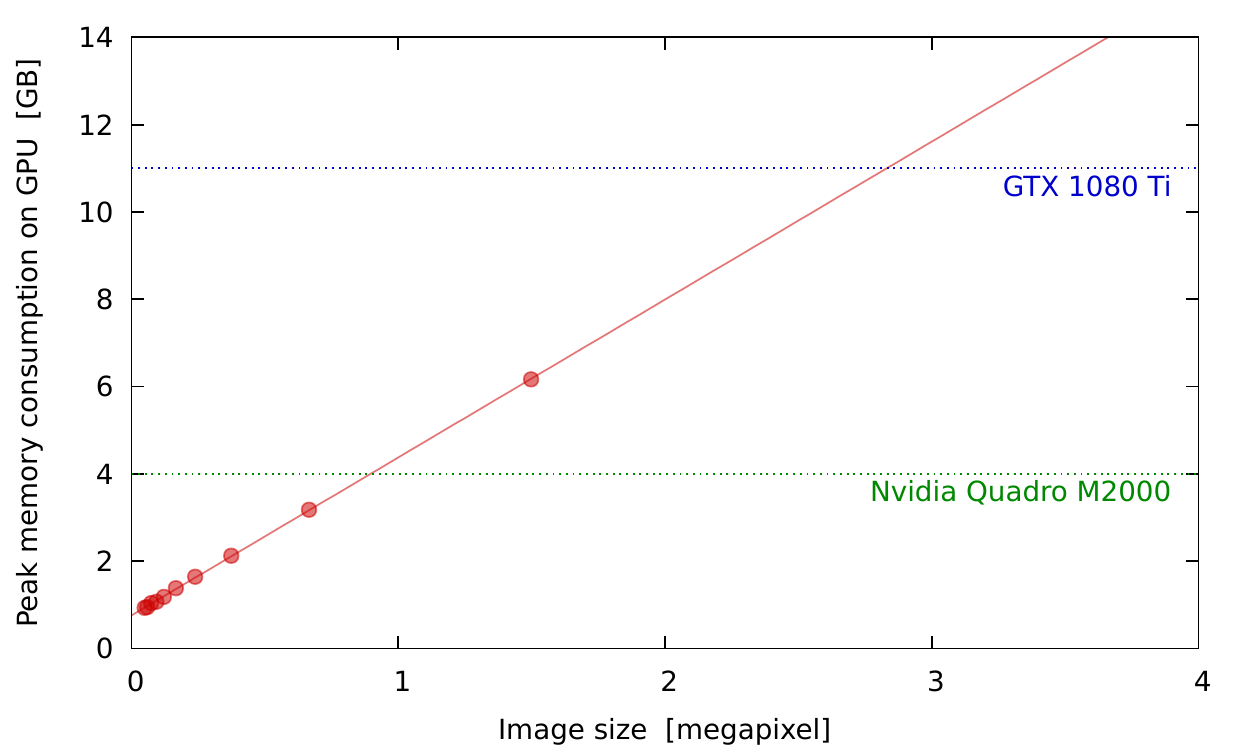}
\caption{%
Observed peak memory consumption of the \textsc{Unit} framework in the 
evaluation phase when translating an image of a certain overall number of 
pixels.
The memory consumption grows roughly linearly with the number of pixels from a 
certain 
threshold.
In addition, we visualize the hardware limits of two GPUs [4GB of an Nvidia 
Quadro M2000 (green) and 11GB of an Nvidia GTX 1080Ti (blue)] as vertical lines.
}
\label{fig:memory_consumption}
\end{figure}
%%%%%%%%%%%%%%%%%%%%%%%%%%%%%%%%%%%%%%%%%%%%%%%%%%%%%%%%%%%%%%%%%%%%%%%%%%%%%%%%

As mentioned in the introduction, there are several approaches in the literature 
to perform domain translation on images.
Each of them has its advantages and shortcomings and we provide a brief 
overview on a selection of methods in this section.
(We refer the reader to the 
respective publications and references therein for details).

One of the very early approaches to unplaired style transfer is the method of 
Gatys~\etal~\cite{gatys2016}.
Their approach is based on a single, pretrained multi-layer CNN which takes the 
image to be transferred as input.
Going deeper and deeper into this CNN, the filters in each layer are activated 
by different image properties, such as \eg~colors, structures, local and global 
features.
Based on the idea that images with a similar style should lead to similar 
activation patterns across the layers of the CNN, they proceed as follows:
Given a target style image and its corresponding activations, the layer 
activations are also determined for the image to be translated.
From the differences in the activations, they determine the gradient 
with respect to the input image and apply this information to change the image.
By this way, the input image resembles more and more to the desired target. 
An advantage is that this method yields high-quality images and that -- by 
setting different gradient weights with respect to different layer depths of 
the CNN -- the style can be adjusted to cover more local or global features.
A drawback of this method, however, is that it requires an optimization 
procedure for each image transformation, which makes this procedure 
computationally very expensive.

A second approach to style transfer of images it the \textsc{Pix2Pix} 
framework by Isola~\etal~\cite{pix2pix}.
Their method is based on a NN in an encoder-decoder configuration whose latent 
space covers the relevant features of the images which are translated.
Compared to the method of Gatys~\etal, the advantage of this approach is that, 
once the network is trained, it can be applied directly to new images without 
additional optimization steps.
This makes the evaluation phase significantly less expensive concerning the 
computational requirements.
However, a shortcoming of this method is that for the network training, 
\emph{paired} images of both domains $\domA$ and $\domB$ are required, which are
often not available in real-world applications.

The much more challenging problem of domain translations on \emph{unpaired} 
image settings has been addressed by the 
\textsc{CycleGAN} \cite{CycleGAN} and 
\textsc{Unit} \cite{UNIT} frameworks.
Both of them are based on extended GAN settings and apply the crucial 
requirement of cycle consistency for training.
In the unpaired setting, a direct transformation from domain $\domA$ to 
domain $\domB$ is not possible, since for an image in $\domA$ there is no 
counterpart in $\domB$.
% This means that, in the unpaired setting, the translation $\domA\to\domB$ 
% (and 
% $\domB\to\domA$) alone is not sufficient because the target image in the 
% second 
% domain, respectively, is not available.
Instead, the transformations $\domA\to\domB\to\domA$ 
and $\domB\to\domA\to\domB$ are performed with the goal to reproduce 
the respective original images.
In both these frameworks, each of the translations $\domA\to\domB$ and 
$\domB\to\domA$ is applied by a generator made of a deep CNN in encoder-decoder 
arrangement, and there are two separate discriminators for each domain $\domA$ 
and $\domB$ which distinguish real and fake images.
The most important difference between the two frameworks is that the 
generators in \textsc{CycleGAN} are completely independent from each other, 
while they share part of their latent space in \textsc{Unit}.
Both these frameworks have shown to yield very good results.
A challenge of these frameworks, however, is the huge overall network size, 
which consists of four different CNNs (two generators and two discriminators; 
see Tab.~\ref{tab:net-config} in the appendix).
This results in a substantial computational effort and long computational times 
for training and evaluation.
In addition, it is required to store all four networks and at least some 
intermediate network results for training on the GPU which sets additional 
hardware requirements.

The latter point is directly related to our main goal of processing 
high-resolution images, so that we go into this in more detail:
We have discovered from our investigations of \eg~the \textsc{Unit} framework, 
that storing an image in the MB range on the GPU is, of course, not 
a problem.
However, propagating the image as float tensors through the network and 
backing up intermediate results for loss functions or gradients for 
backpropagation, the peak memory consumption of the whole network is 
significantly higher, with an especially large contribution from the deep 
hidden layers.
We illustrate this observation in \FIG~\ref{fig:memory_consumption}, which 
shows from a threshold a linear increase of the peak memory consumption with 
the input image's number of pixels.
From the vertical lines, which indicate the GPU memory hardware limit of two 
GPUs, it becomes clear that this framework cannot be executed \eg~on a standard 
Nvidia Quadro M2000 for images with more that about one megapixel and the limit 
on a Nvidia GTX 1080Ti is less than three megapixel.
Even if code improvements might reduce the memory consumption, this basic 
limit to the processable image resolution remains, and also special high-prize 
computing GPUs which provide higher GPU memory can only push this boundary 
but cannot overcome it.

%%%%%%%%%%%%%%%%%%%%%%%%%%%%%%%%%%%%%%%%%%%%%%%%%%%%%%%%%%%%%%%%%%%%%%%%%%%%%%%%
\begin{figure}[t]
\includegraphics[width=\columnwidth]{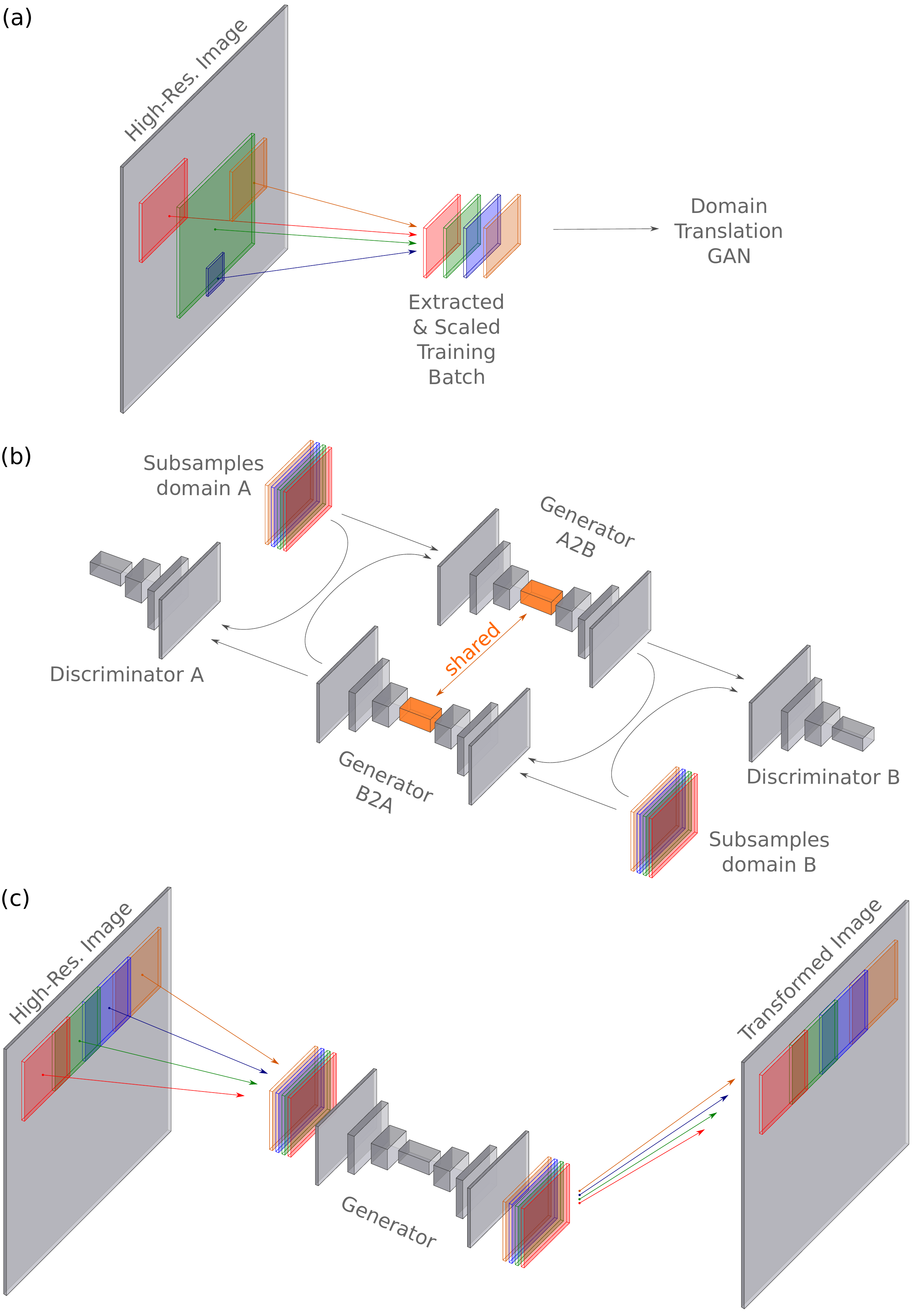}
\caption{%
(a) 
From the high-resolution image, training samples are extracted from 
random positions and with random size.
The extracted samples are scaled down to a common resolution 
$x_\text{batch} \times y_\text{batch}$ and combined to a training batch.
(b)
With this extracted batch, a training iteration of the GAN setting is performed.
For the GAN, known configurations like the \textsc{CycleGAN} \cite{CycleGAN} or 
\textsc{Unit} \cite{UNIT} frameworks can be used and we use the latter with 
shared latent space in this paper.
(c)
To transform the whole high-resolution image after training, single samples are 
extracted and translated by the generator separately.
The translated images are finally merged together.
We emphasize that in the translation step, the single samples may overlap and 
also be of different size.
}
\label{fig:method}
\end{figure}

%%%%%%%%%%%%%%%%%%%%%%%%%%%%%%%%%%%%%%%%%%%%%%%%%%%%%%%%%%%%%%%%%%%%%%%%%%%%%%%%

\section{Method}
\label{sec:method}

%%%%%%%%%%%%%%%%%%%%%%%%%%%%%%%%%%%%%%%%%%%%%%%%%%%%%%%%%%%%%%%%%%%%%%%%%%%%%%%%
\begin{figure*}[p]
% \includegraphics[width=\textwidth]%
% {fig/highres/landscape2/landscape2_ori_3000.jpg} \\
% \includegraphics[width=\textwidth]%
% {fig/highres/landscape2/landscape2_trans_3000.jpg} \\
% \textcolor{white}{x}\hfill
% \includegraphics[width=.14\textwidth]%
% {fig/highres/landscape2/landscape2_style_1000.jpg} 
\includegraphics[width=\textwidth]{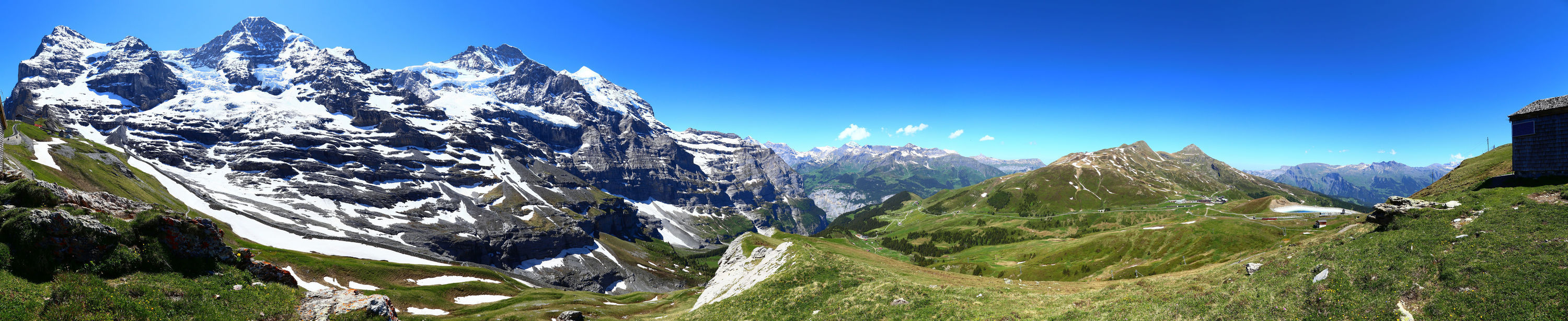} \\
\includegraphics[width=\textwidth]{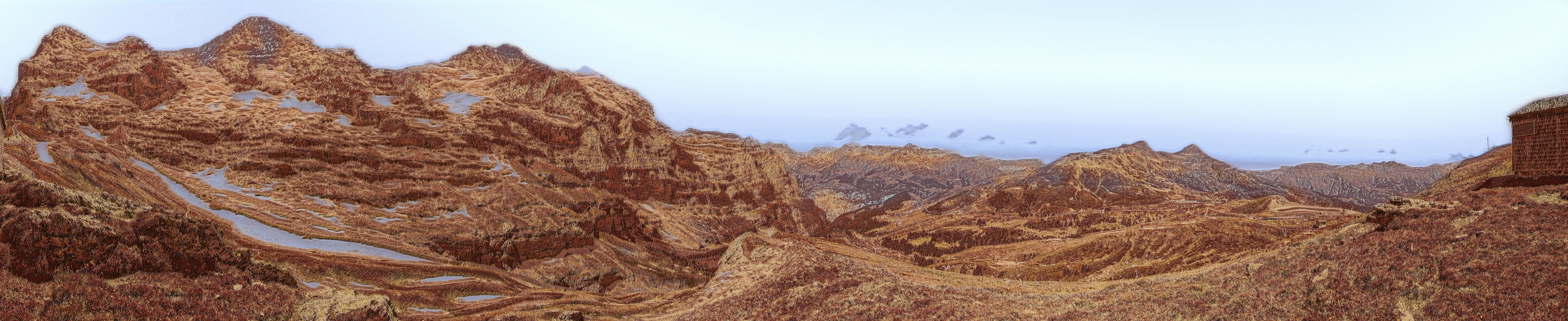} \\
\textcolor{white}{x}\hfill
\includegraphics[width=.14\textwidth]{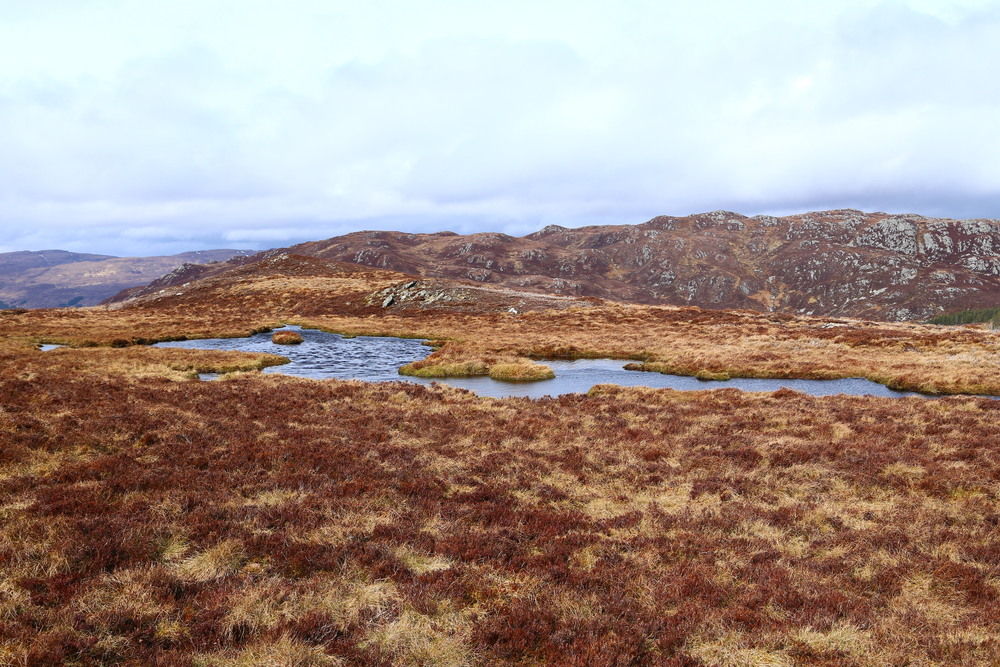} 
\caption{%
High-resolution style transfer of an image of the Swiss Alps (top) towards the 
style of the Scottish Highlands (bottom). 
Domain $\domA$ is defined by the top image and the target style is defined by 
the small image on the bottom right.
The original image has a resolution of $15,884 \times 3,271$ pixels and the
target style image of $5,472 \times 3,648$ pixels.
(The resolution has been downscaled for the presentation in this paper.)
Image details can be seen in the two top rows of 
\FIG~\ref{fig:landscape2_details}.
}
\label{fig:landscape2}
\vspace{3em}
% 
% \includegraphics[width=\textwidth]%
% {fig/highres/street2/street2_ori_3000.jpg} \\
% \includegraphics[width=\textwidth]%
% {fig/highres/street2/street2_trans_3000.jpg} \\
% \textcolor{white}{x}\hfill
% \includegraphics[width=.14\textwidth]%
% {fig/highres/street2/street2_style_1000.jpg}
\includegraphics[width=\textwidth]{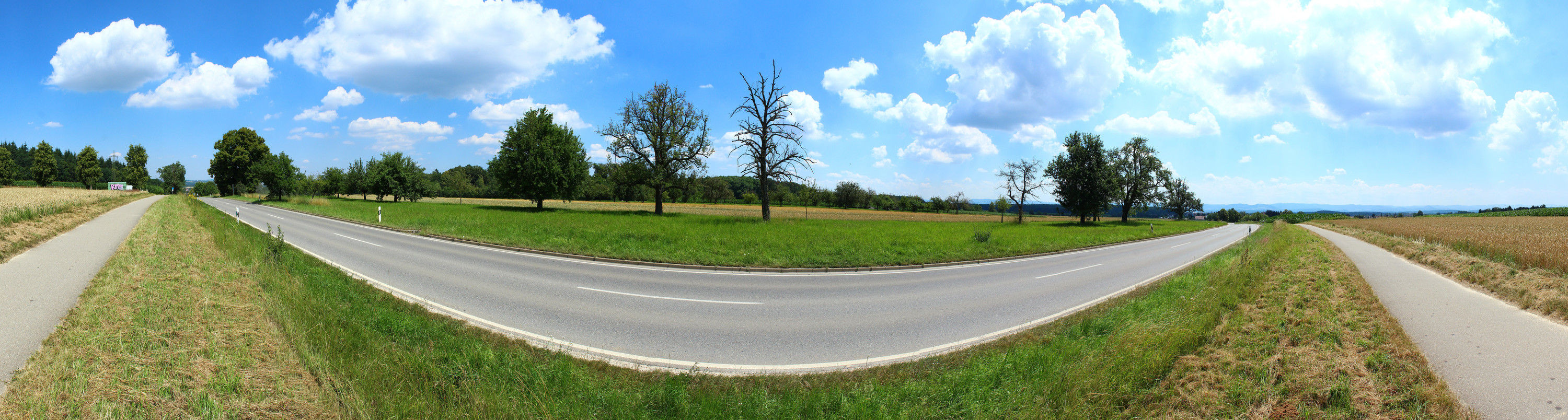} \\
\includegraphics[width=\textwidth]{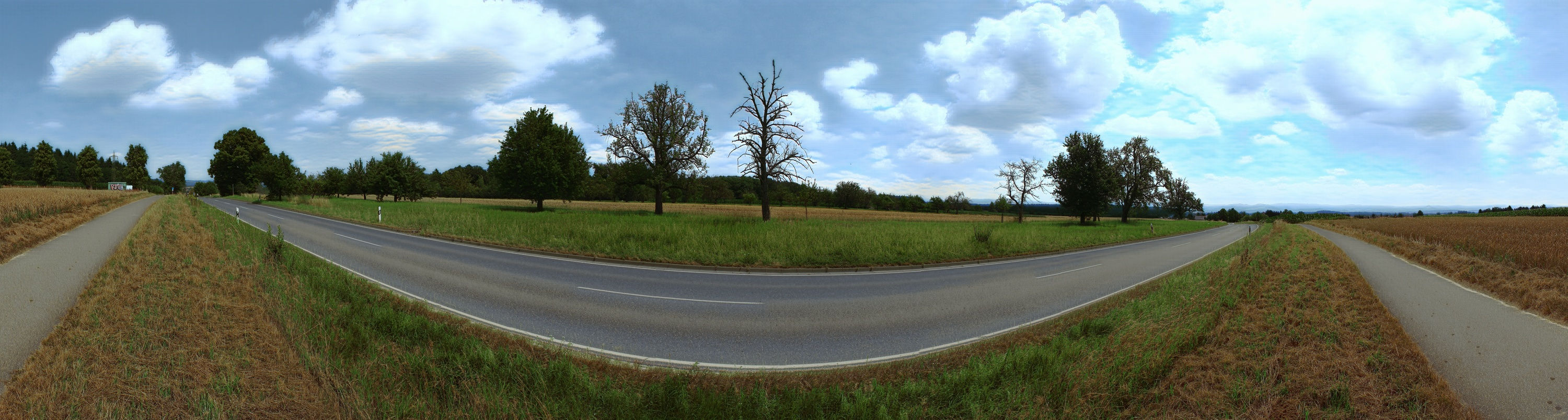} \\
\textcolor{white}{x}\hfill
\includegraphics[width=.14\textwidth]{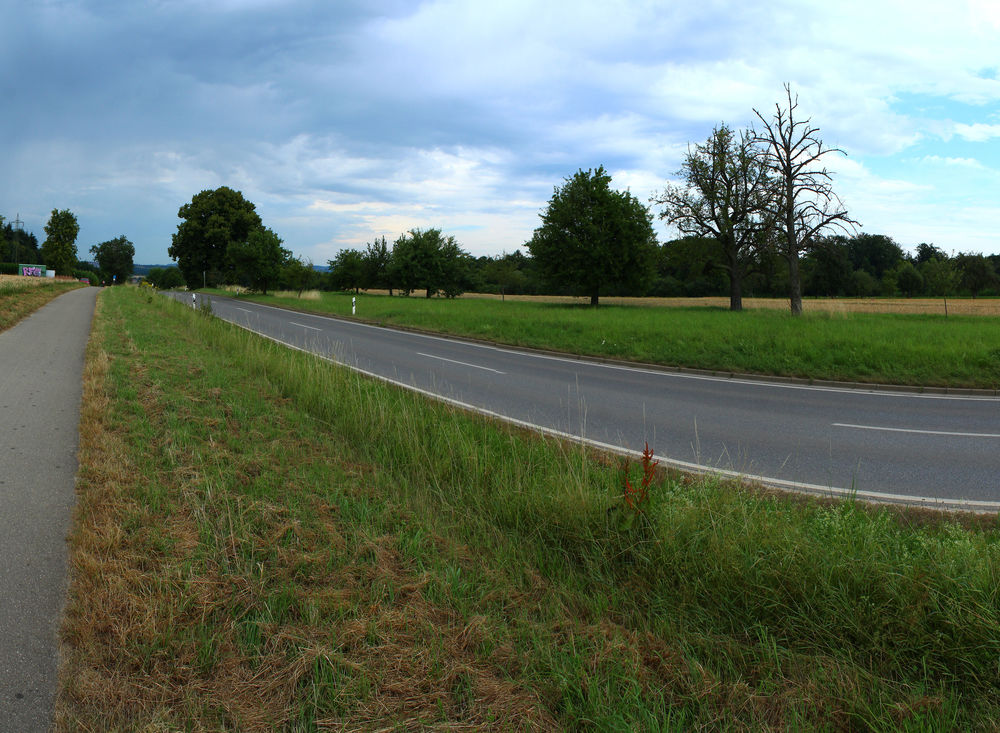}
\caption{%
High-resolution style transfer of a street scene between very similar-looking 
domains.
The original image (top) and the target style image (bottom right) show the 
same street, but with images taken from different positions and under different 
weather and lighting conditions.
The transformed image (center) well exhibits the style of the target 
domain.
(The original image has a size of $12,895 \times 3,472$ pixels and the target 
image one of $4,785 \times 3,508$ pixels.)
Image details can be seen in \FIG~\ref{fig:landscape2_details}.
}
\label{fig:street1}
\end{figure*}
%%%%%%%%%%%%%%%%%%%%%%%%%%%%%%%%%%%%%%%%%%%%%%%%%%%%%%%%%%%%%%%%%%%%%%%%%%%%%%%%

It is the purpose of this paper to introduce an approach by which 
arbitrary-high resolution images can be processed on today's standard GPUs.
The basic idea is simple and can be stated in one sentence:
Instead of processing the whole image at once, perform the network training as 
well as its evaluation on small subsamples of the image.
From a mathematical point of view, the justification of this procedure is the 
analogous functional principle of the CNN's filters which stride over the input 
image:
With a usual size of three to seven pixels, these filters are very small 
compared to the actual image and they only see a very small part of it in every 
step.
Our procedure of extracting subsamples can, therefore, be regarded as an 
abstract intermediate interface to the NN which works on a level between 
the whole image size and the filter size.
In more detail, the procedure is described in the following two subsections 
and illustrated in \FIG~\ref{fig:method}.

\subsection{Training}

To train an unpaired domain translation network for high-resolution 
images, we start from a training set consisting of a single or several such 
images and extract a batch out of the high-resolution image as shown in 
\FIG~\ref{fig:method}(a):
According to the desired batch size $b$, we extract image subsamples of 
different size out of the original image.
Both, position and size of each subsample can be chosen arbitrarily and all 
extracted samples are then scaled down to a common resolution
\begin{equation}
 x_\text{batch} \times y_\text{batch} \,.
\end{equation}
Including $n_\text{color}$ color channels, the respective batch tensor then has 
the size
\begin{equation}
 b \otimes x_\text{batch} \otimes y_\text{batch} \otimes n_\text{color} \,.
\end{equation}
With this tensor, we perform a single training iteration of the underlying 
unpaired domain translation network as illustrated in \FIG~\ref{fig:method}(b) 
(\eg~a \textsc{CycleGAN} \cite{CycleGAN} or \textsc{Unit} \cite{UNIT} 
framework).
This step of sample extraction and update iteration is repeated until the 
algorithm has converged or another stopping criterion is reached.

\subsection{Translation}

Analogously to the training procedure, the evaluation phase is performed 
on the level of small image subsamples.
As illustrated in \FIG~\ref{fig:method}(c), in a first step samples are 
extracted from the image that shall be transformed.
Second, each of them is translated to the other domain by the generator 
separately.
Finally, the translated images are
merged together to the translated high-resolution image.

\subsection{Remarks}

The mechanism to train and evaluate NNs for the domain 
translation task described above leaves some freedom in the actual application. 
Therefore, we want to extend this scheme by some remarks:

First, for high-resolution images of pixel size $x_\text{full} \times 
y_\text{full}$ and extracted small batches of size $x_\text{batch} \times 
y_\text{batch}$, the number of possible, different image subsamples is
\begin{equation}
 (x_\text{full} - x_\text{batch}) \times (y_\text{full} - y_\text{batch}) \gg 1 
\,.
\label{eq:number_of_possibilities}
\end{equation}
Taking as an example an image of size $5000 \times 3000$ pixels and working 
on subsamples of size $128 \times 128$, the value on the left-hand side of 
\EQ~\eqref{eq:number_of_possibilities} is almost 14 million.
Further taking into account the different sizes of the extracted samples and 
possible horizontal or vertical image flips, this number becomes even larger.
This makes clear that, by this procedure, one single high-resolution image 
can effectively act as a huge training set.
Using random positions and sizes, it is also very unlikely that the NN sees 
exactly the same image more than once during the whole training process, which 
prevents overfitting.

Second, in the extraction phase during training, we use random positions over 
the whole image and random sizes in the range between the small batch 
($x_\text{batch} \times y_\text{batch}$) 
and the whole image 
($x_\text{full} \times y_\text{full}$).
This corresponds to different zoom levels of the image and, hence, we 
cover different length scales of the image as well as help the generator to 
learn, both, global and local image properties.
The different zoom levels of the exracted images can also be interpreted as 
effectively changing the distance from camera to object which helps the 
generator to generalize better along the optical axis of the camera in 
addition to the axes perpendicular to it.

Third, an advantage of only processing small subsamples
is that the peak memory consumption  
during the training and evaluation phase is set by the subsample size 
$x_\text{batch} \times y_\text{batch}$, but not by the size $x_\text{full} 
\times y_\text{full}$ of the full high-resolution image. 
Larger images ``only'' lead to larger computation times during the evaluation 
phase, because more subsamples need to be processed, but they do not increase 
the GPU memory requirements.
Nevertheless, it is, of course, possible to fully parallellize the processing 
of the image subsamples if more computational resources are available.

Each training batch can be extracted from only one or, of course, also from 
several independent images.

In the translation phase, we use overlapping subsamples and use the average 
color value for each pixel from the different samples.
Our experiments have shown that this improves image quality by reducing noise 
and avoiding that neighboring samples show discontinuities of objects at their 
borders.

Let us finally remark that, due to the evaluation level on subsamples, 
images sizes in the two domains can be chosen independently.
It is also possible to train models on different image sizes than those which 
they are evaluated on later.
Of course, a trained generator can be applied to further images.

\section{Results and Discussion}
\label{sec:results}

%%%%%%%%%%%%%%%%%%%%%%%%%%%%%%%%%%%%%%%%%%%%%%%%%%%%%%%%%%%%%%%%%%%%%%%%%%%%%%%%
\begin{figure*}[t]
% \includegraphics[width=.499\textwidth]%
% {fig/highres/street3/street3_ori_2000.jpg} 
% \includegraphics[width=.499\textwidth]%
% {fig/highres/street3/street3_trans_2000.jpg} \\
% \textcolor{white}{x}\hfill
% \includegraphics[width=.15\textwidth]%
% {fig/highres/street3/street3_style_1000.jpg}
\includegraphics[width=.499\textwidth]{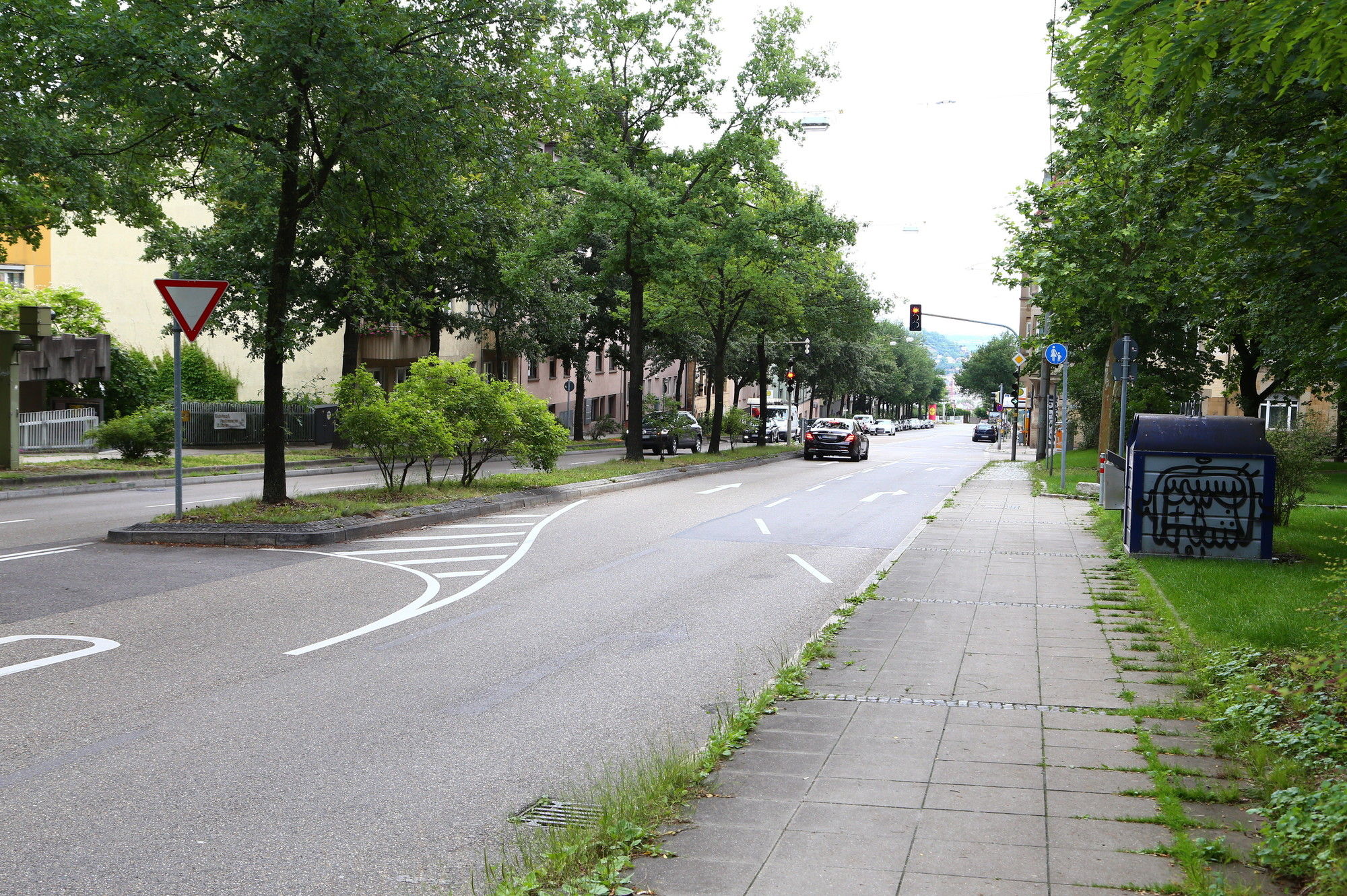} 
\includegraphics[width=.499\textwidth]{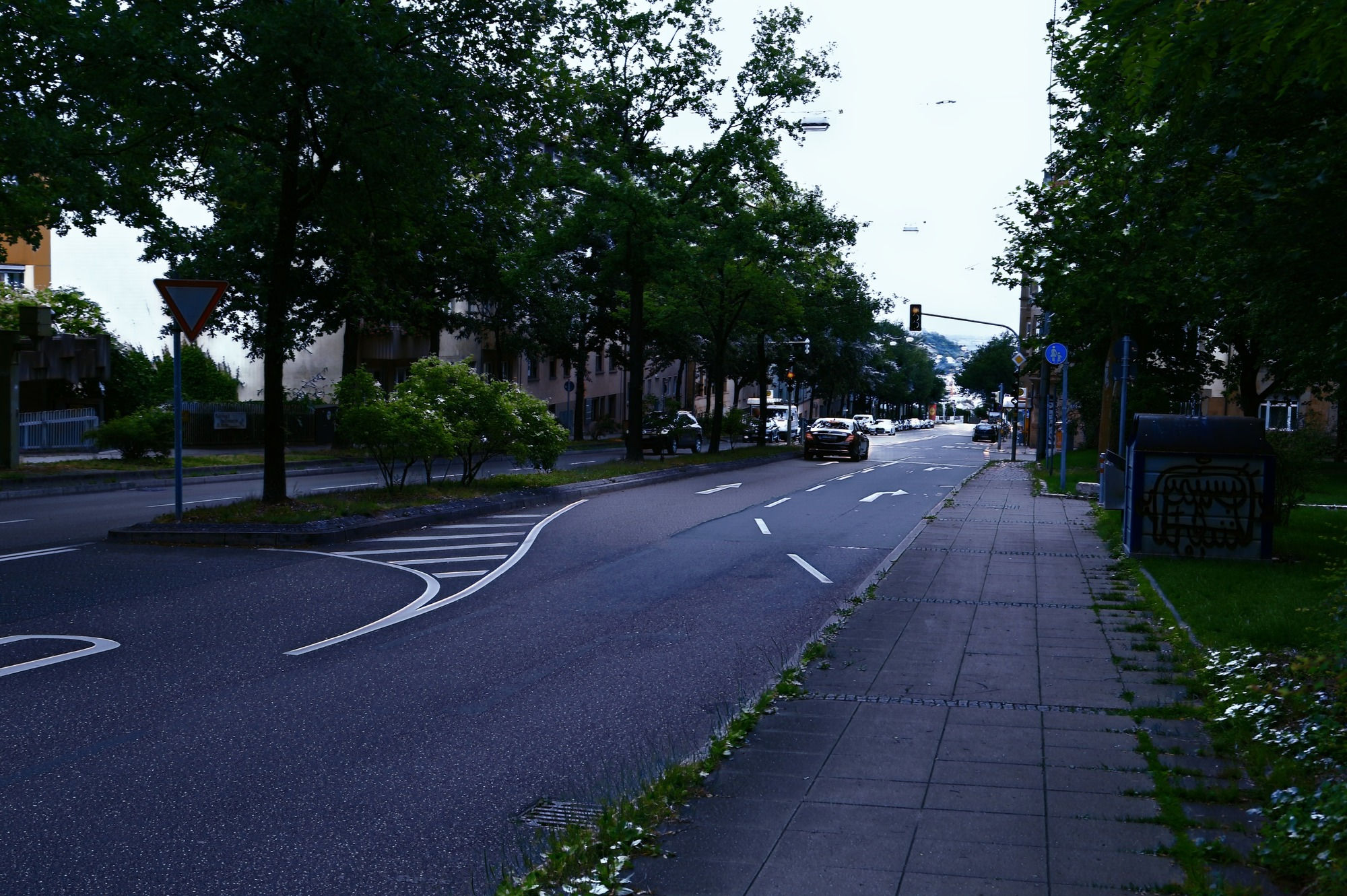} \\
\textcolor{white}{x}\hfill
\includegraphics[width=.15\textwidth]{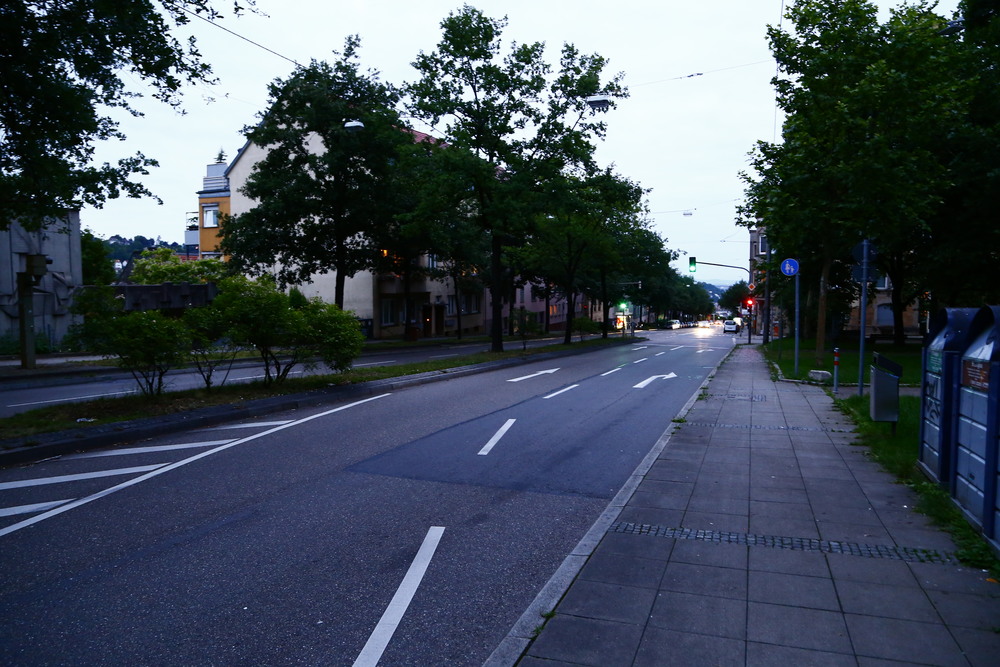}
\caption{%
Domain translation of a high-resolution street scene from sunny to darker 
lighting conditions.
In this case, original and target image are of the same size of $5,472 
\times 3,648$ pixels. 
Image details can be seen in \FIG~\ref{fig:landscape2_details}.
}
\label{fig:street2}
\end{figure*}
%%%%%%%%%%%%%%%%%%%%%%%%%%%%%%%%%%%%%%%%%%%%%%%%%%%%%%%%%%%%%%%%%%%%%%%%%%%%%%%%

In this section, we present results of the domain translation obtained with 
different high-resolution image sizes and styles.
In all cases, the original and transformed images are too large to be properly 
presented in a pdf document, so that we have downscaled them to a reasonable 
file size.
To provide a detailed view on some images, we present characteristic image 
details in the appendix.

Concerning the style of the images, we present different domain adaption 
ranges, meaning that in some images, the domains $\domA$ and $\domB$ are very 
different and in some others they are very similar.
All domain translations in this paper have been performed in a low-variance 
``one-shot'' setting, in which both domains are defined by only one 
high-resolution image.

All images in this paper have been translated by the procedure described in 
\SEC~\ref{sec:method} with subsample size of \mbox{$128 \times 128$} pixels.
Moreover, a \textsc{Unit}-like GAN-setting \cite{UNIT} with shared latent space 
and a configuration as listed in Tab.~\ref{tab:net-config} in the appendix has 
been used.
In order to underline the small memory consumption of the procedure, we 
emphasize that we have trained and evaluated all the NNs for the images shown 
in this paper on a ``small'' standard Desktop GPU Nvidia Quadro M2000 with only 
4GB GPU memory.
% Cross-checking the images with results obtained on an Nvidia GTX 1080Ti 
% confirmed that there is no lack of quality related to limited hardware.

As a first example to demonstrate the performance of the presented method, we 
show in \FIG~\ref{fig:landscape2} the domain translation of a panorama image 
taken in the Swiss Alps towards the style of the Scottish Highlands, which are 
very different-looking image domains.
The original image has a resolution of \mbox{$15,884 \times 3,271$} pixels 
(more than 50 megapixels altogether) 
and the target style is defined by an image with \mbox{$5,472 \times 3,648$}
pixels (about 20 megapixels).
We see that the target style is well adopted by the translation on both 
local and global length scales.
The clear blue sky is transformed throughout to a cloudy one, and grass as well 
as rocks receive the style of the brownish Scottish Highland landscape.
Interestingly, only some snow fields are interpreted as water while others also 
are translated to brown earth, which nicely demonstrates that the NN not only 
repaints the different areas but takes into account its surrounding and meaning.
Also, the image details are very well preserved as we show by the subsamples 
presented in the appendix (see \FIG~\ref{fig:landscape2_details}).
Even single trees, small paths, and also blades of grass keep their structure.

A second panorama image style transfer is presented in \FIG~\ref{fig:street1} 
showing a street scenery.
With this example, we address the situation of very similar domains $\domA$ and 
$\domB$.
Both the original as well as the target image show the same street, but the 
images have been taken from different positions and under different weather and 
lighting conditions.
The original image is of size 
\mbox{$12,895 \times 3,472$} pixels (about 45 megapixels) and the target image 
has a resolution of \mbox{$4,785 \times 3,508$} pixels.
Also in this case of similar styles, the procedure performs well, and it 
preserves, both, local and global image information.
Image details are, again, provided in \FIG~\ref{fig:landscape2_details} in the 
appendix.

Figure~\ref{fig:street2} shows a street scene as well, but with more complex 
image 
content.
Beyond the street, grass and trees, this image contains also a sidewalk, more 
complex road markings, a car, traffic lights and signs, and some houses in the 
background.
Here, both the original and the target image have the same resolution of 
\mbox{$5,472 \times 3,648$} pixels (about 20 megapixels). 
Again, we focus on the case of rather similar domains and take as target an 
image showing the same street, but photographed shortly before sunset and from 
another position, whereas the source was shot in bright daylight.
Again, the image is well transformed to the target style and 
contains all details which define the scene.
(See \FIG~\ref{fig:landscape2_details} in the appendix for image details.)

In \FIG~\ref{fig:street5}, we demonstrate once more that our method is capable 
of dealing with larger domain differences. 
Therefore, we show the cross-translation of a street and a dirt road, both 
images having a 
size of \mbox{$5,472 \times 3,648$} pixels (about 20 megapixels).
The top row of this figure shows the original photographs and the bottom row 
shows the translated images in the domain that is defined by the respective 
other photograph.
The generator has learned to transform the asphalt street into a 
dirty and stony surface, while the grass structure and the sky keep their 
structure while being only translated in color.
(See \FIG~\ref{fig:landscape2_details} in the appendix for image details.)

%%%%%%%%%%%%%%%%%%%%%%%%%%%%%%%%%%%%%%%%%%%%%%%%%%%%%%%%%%%%%%%%%%%%%%%%%%%%%%%%
\begin{figure*}[p]
% 
% \includegraphics[width=.49\textwidth]%
% {fig/highres/street5/street5A_ori_1000.jpg} \hfill
% \includegraphics[width=.49\textwidth]%
% {fig/highres/street5/street5B_ori_1000.jpg} \\[1ex]
% \includegraphics[width=.49\textwidth]%
% {fig/highres/street5/street5A_trans_1000.jpg} \hfill
% \includegraphics[width=.49\textwidth]%
% {fig/highres/street5/street5B_trans_1000.jpg} 
\includegraphics[width=.49\textwidth]{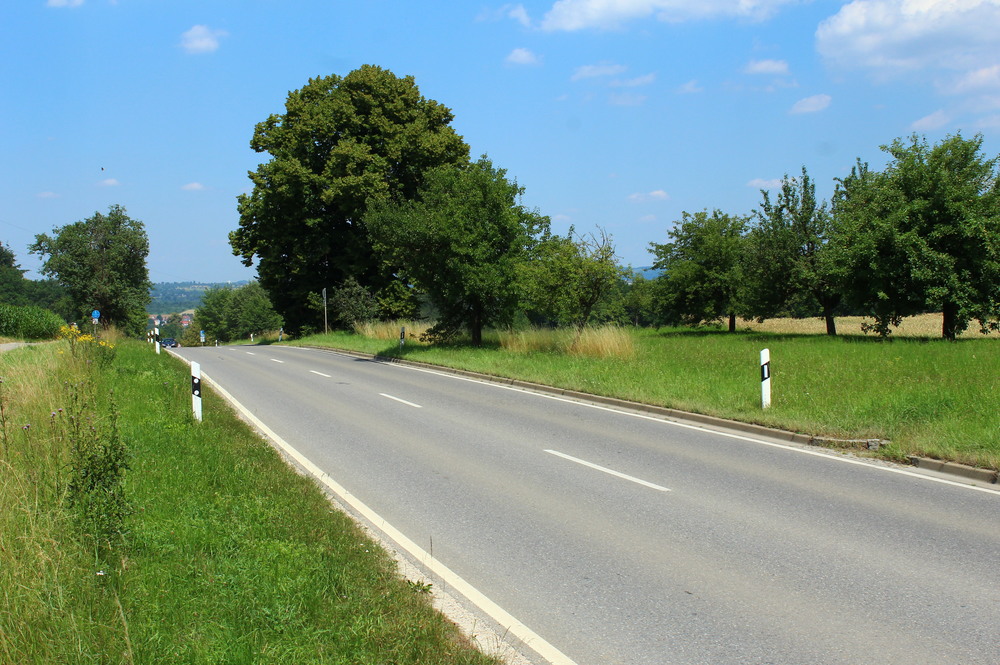} \hfill
\includegraphics[width=.49\textwidth]{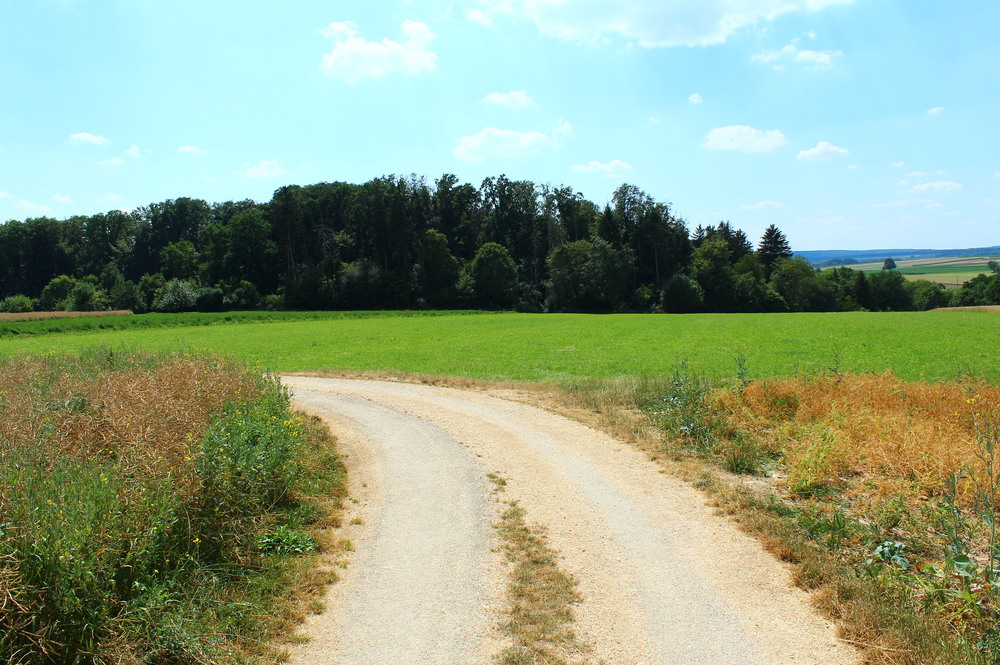} \\[1ex]
\includegraphics[width=.49\textwidth]{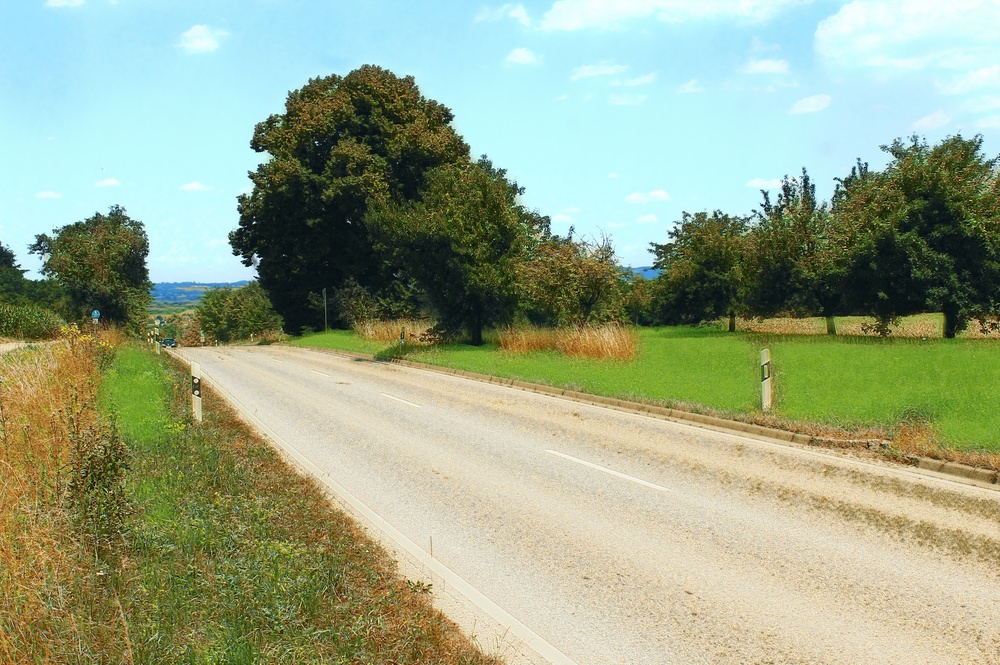} \hfill
\includegraphics[width=.49\textwidth]{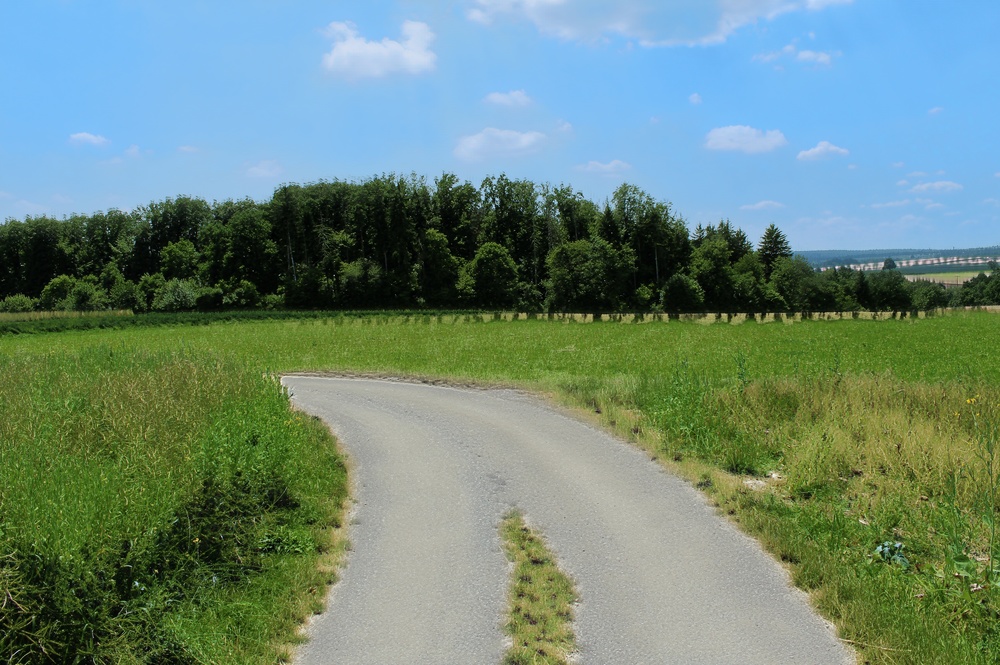} 
\caption{%
Cross-domain-translation between a high-resolution street image and one of 
a dirt road.
Both images are of the same size of $5,472 \times 3,648$ pixels and the target 
domain is the respective other scene.
Image details can be seen in the two center rows of 
\FIG~\ref{fig:landscape2_details}.
}
\label{fig:street5}
\vspace{3em}
% 
% \includegraphics[width=.47\columnwidth]%
% {fig/highres/trafficsign_1/trafficsign1_ori_1000.jpg}
% \includegraphics[width=.47\columnwidth]%
% {fig/highres/trafficsign_3/trafficsign3_ori_1000.jpg}
% \hfill
% \includegraphics[width=.47\columnwidth]%
% {fig/highres/trafficsign_2/trafficsign2_ori_1000.jpg}
% \includegraphics[width=.47\columnwidth]%
% {fig/highres/trafficsign_4/trafficsign4_ori_1000.jpg}
% \\
% \includegraphics[width=.47\columnwidth]%
% {fig/highres/trafficsign_1/trafficsign1_trans_1000.jpg}
% \includegraphics[width=.47\columnwidth]%
% {fig/highres/trafficsign_3/trafficsign3_trans_1000.jpg}
% \hfill
% \includegraphics[width=.47\columnwidth]%
% {fig/highres/trafficsign_2/trafficsign2_trans_1000.jpg}
% \includegraphics[width=.47\columnwidth]%
% {fig/highres/trafficsign_4/trafficsign4_trans_1000.jpg}
\includegraphics[width=.47\columnwidth]{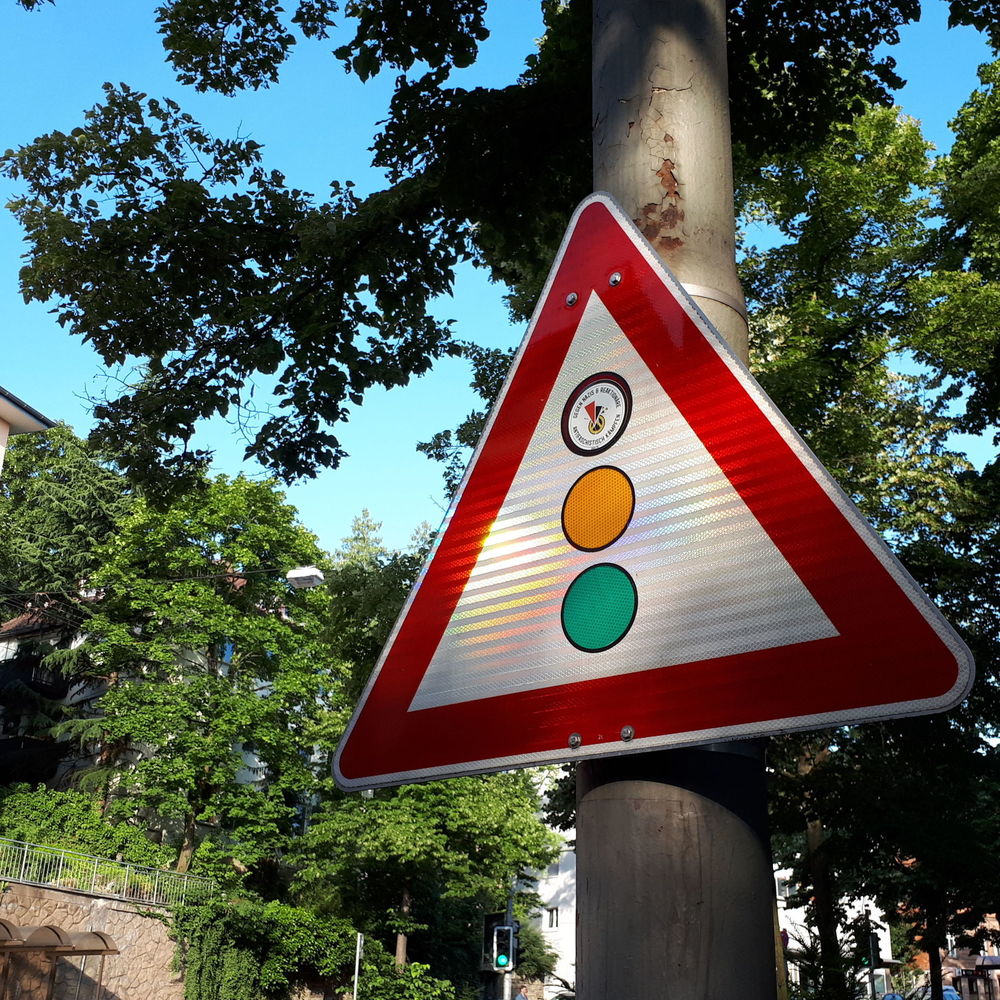}
\includegraphics[width=.47\columnwidth]{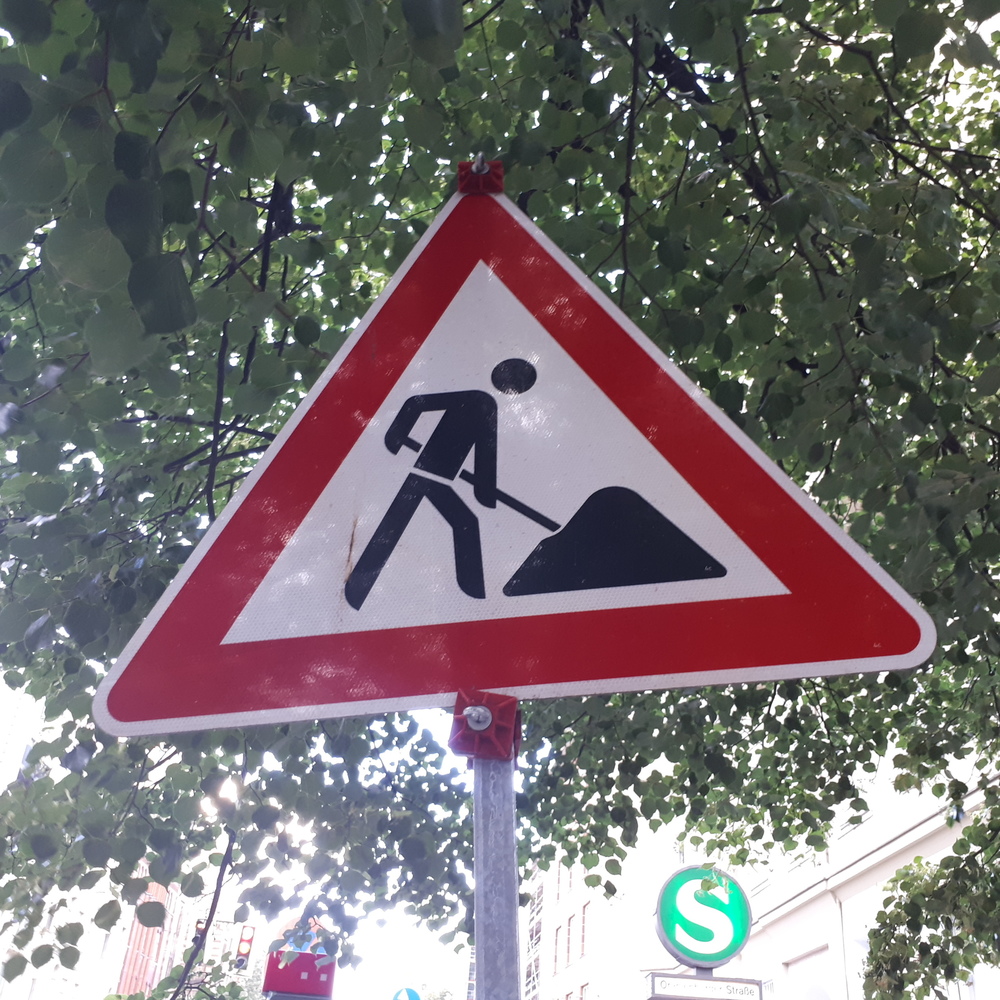}
\hfill
\includegraphics[width=.47\columnwidth]{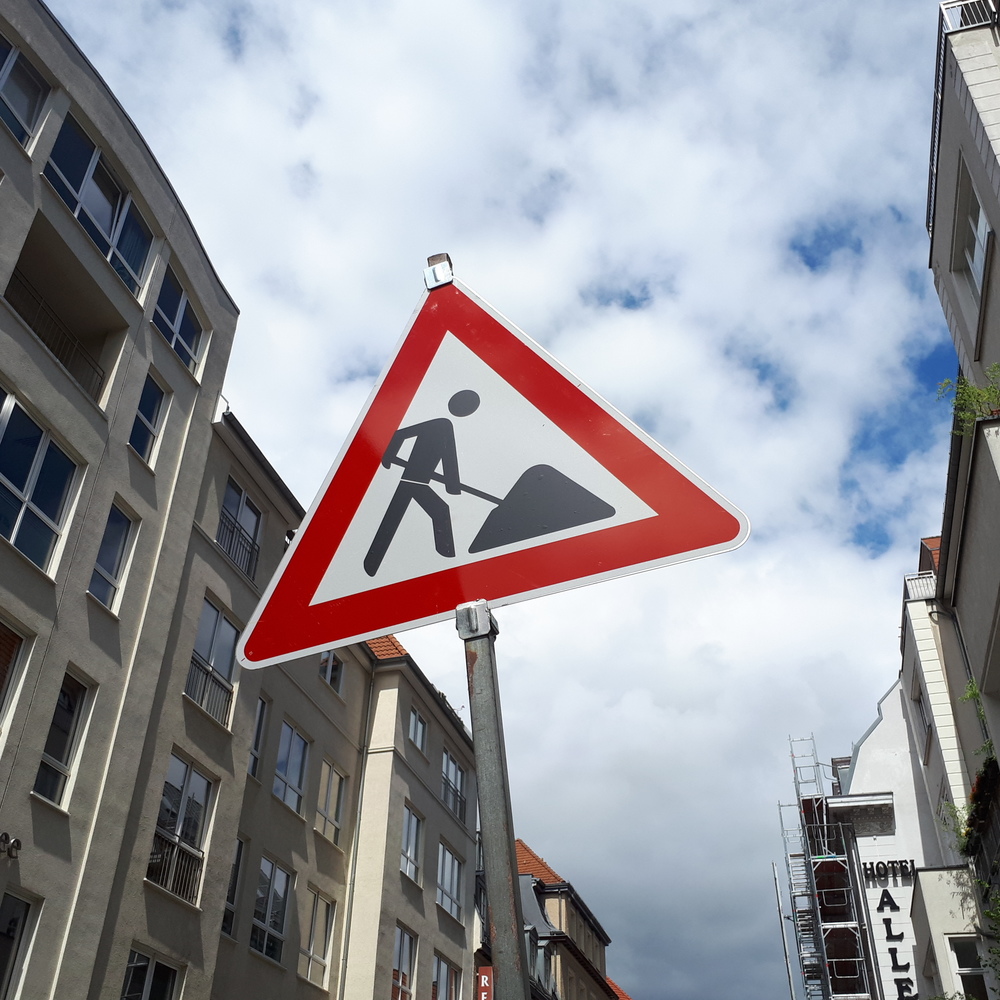}
\includegraphics[width=.47\columnwidth]{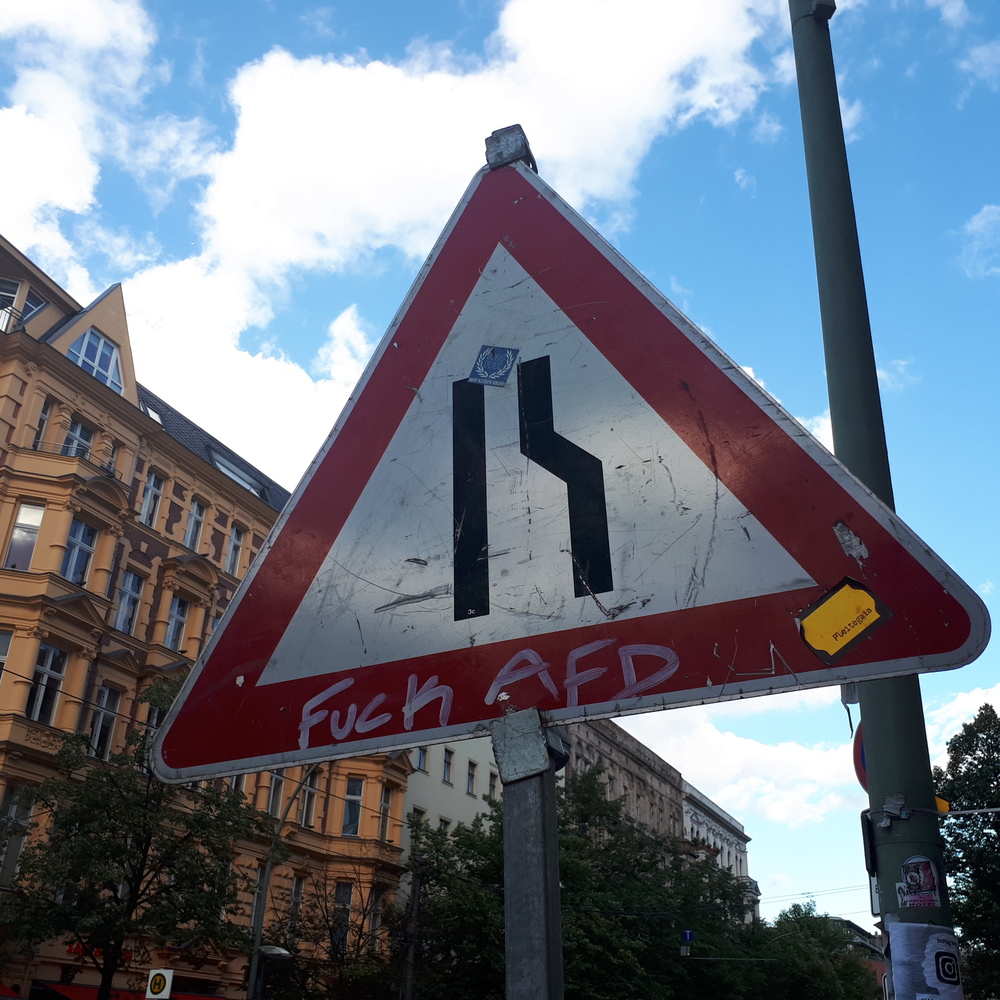}
\\
\includegraphics[width=.47\columnwidth]{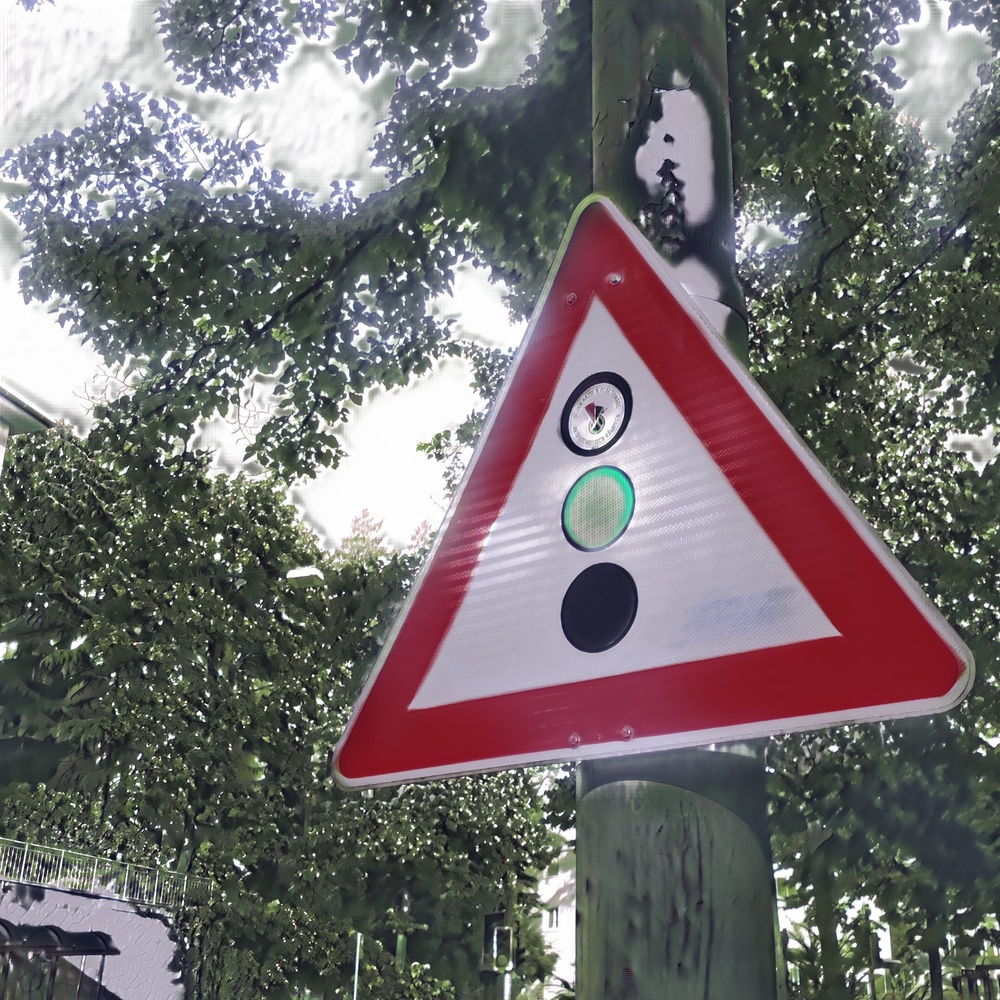}
\includegraphics[width=.47\columnwidth]{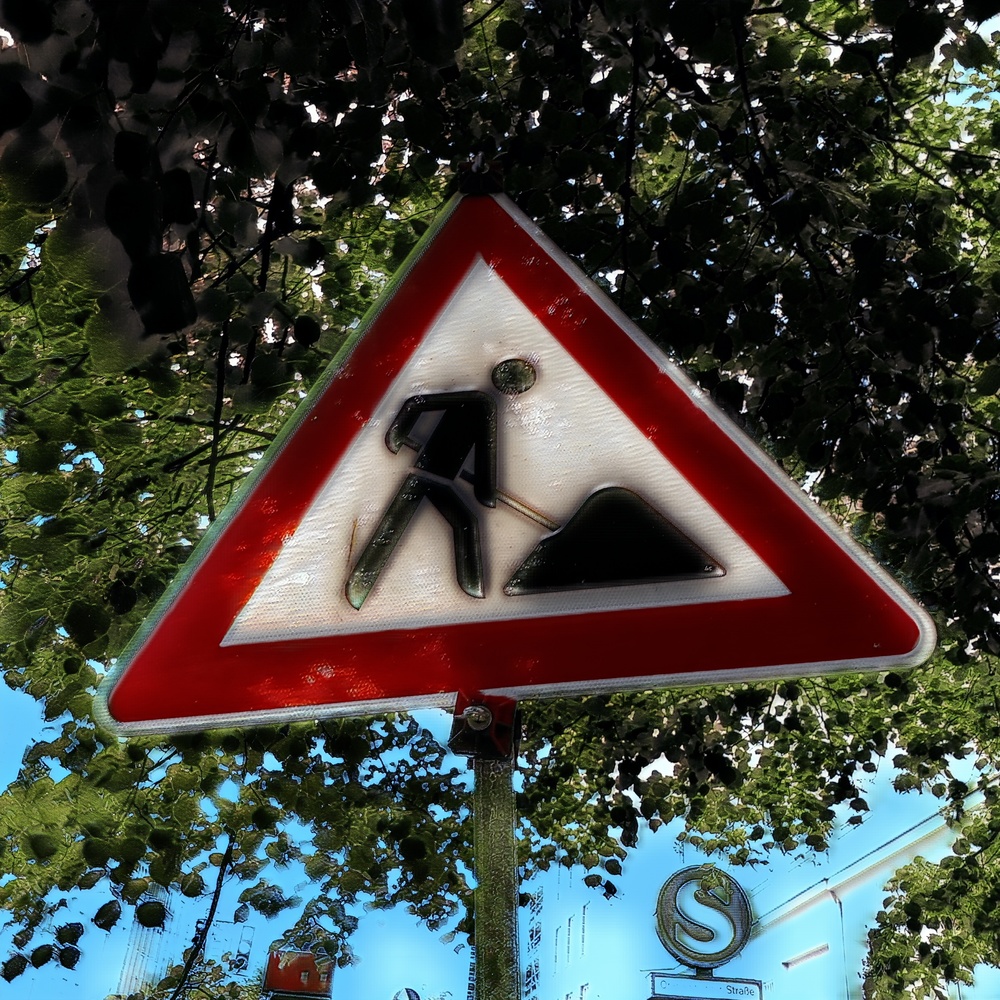}
\hfill
\includegraphics[width=.47\columnwidth]{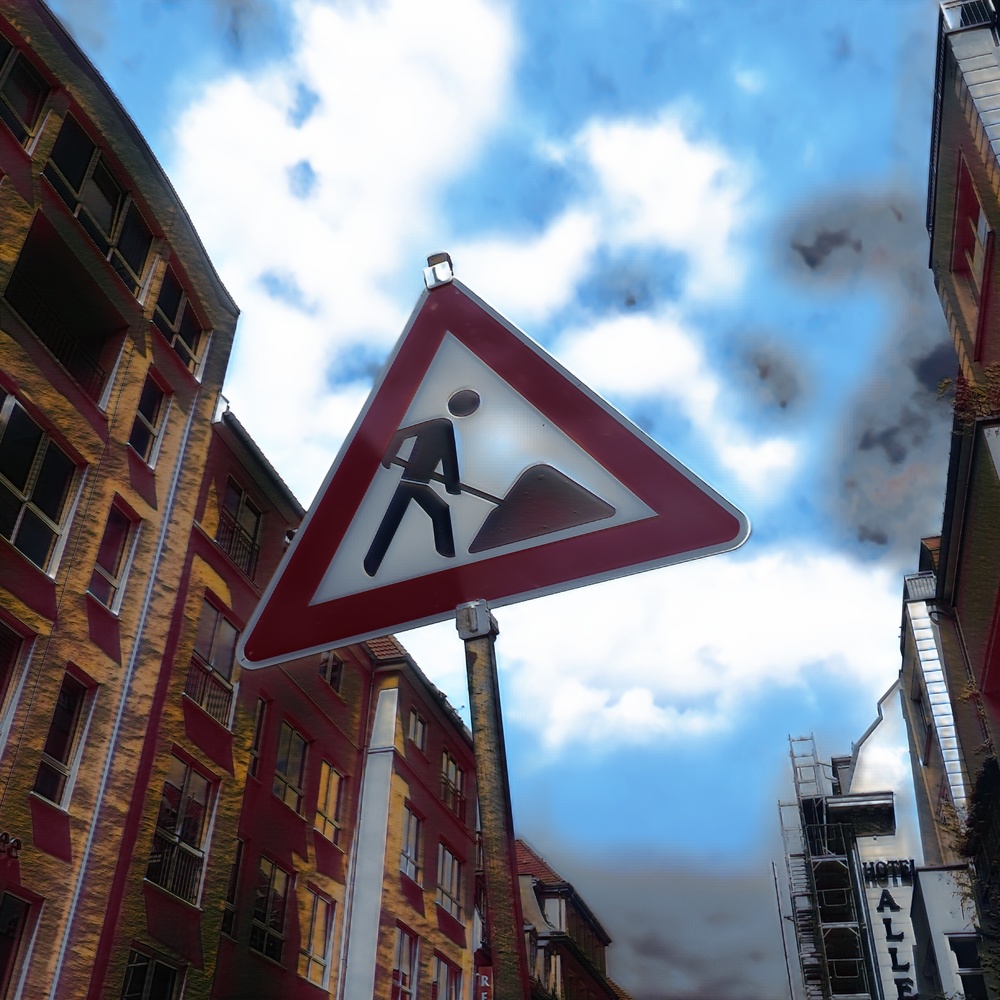}
\includegraphics[width=.47\columnwidth]{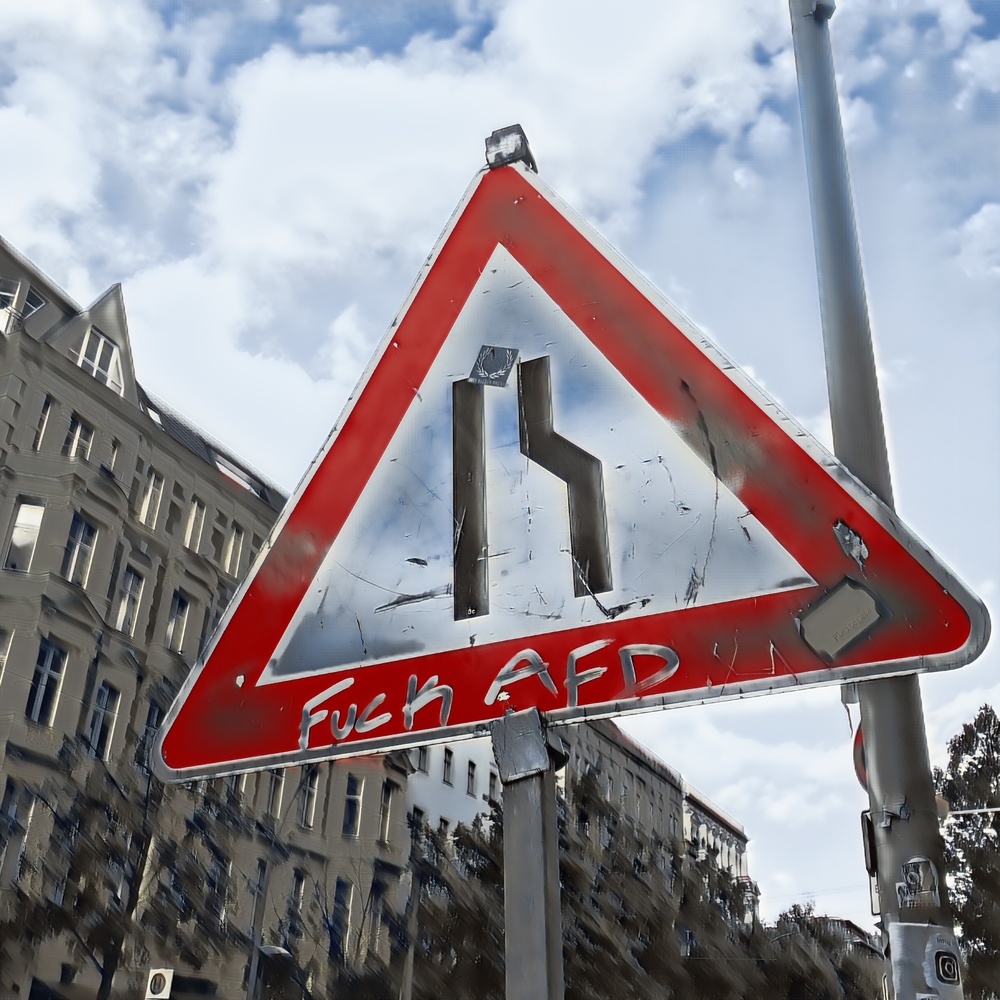}
\caption{%
High-resolution domain transfer of traffic signs:
The four images on the left belong to one translation process, and the four on 
the right to another.
The top row shows the original images and the bottom row the translated ones.
In this figure, we show a cross-translation, i.\,e.~the images are pairwise 
translated to the domain which is defined by the other image.
(The translations have been performed on images with resolution $2,448 \times 
2,448$ pixels.)
}
\label{fig:trafficsign}
\end{figure*}
%%%%%%%%%%%%%%%%%%%%%%%%%%%%%%%%%%%%%%%%%%%%%%%%%%%%%%%%%%%%%%%%%%%%%%%%%%%%%%%%

We finish the results focussing on cross-translations of traffic signs in 
\FIG~\ref{fig:trafficsign}.
Here, we show two groups of traffic sign images, respectively, on the left and 
on the right.
The top row shows the original images and the bottom row the translated ones.
Here, the images in the top row are assigned pairwise to the domains $\domA$ 
and $\domB$ and in the bottom row, the same traffic signs are shown in the 
domain, which is defined by the respective other image.
Each of the translations has been performed on images of size 
\mbox{$2,448 \times 2,448$} pixels (about 6 megapixel). 
In all the cases, the structure of the images and the content is well 
translated.
The only exception is that the colors yellow and green of the traffic light 
symbol are not preserved (left column).
However, this is clear since the color yellow does not occur in the target 
domain (second column).
From this perspective, this observation is in agreement with the goal to reach 
the target domain, but it also makes clear how important the choice of the 
target data set is for the translation task.

% Beyond the experiments presented above, we have performed style transfers also 
% for further domains, such as \eg~the task to translate simulated into 
% photo-realistic images or that of translating video style by applying the 
% procedure to the single video frames (both not shown in this paper).
% Also in these cases, the procedure performed well and yielded high-quality 
% images.
% In addition, we made the observation that for videos this procedure helps to 
% improve the temporal image consistency.
We note that, despite the huge resolution that has finally been processed, 
training time of the whole GAN setting was on the order of only one day even on 
the small Nvidia Quadro GPU.
One of our general observations is also that using smaller image subsamples 
helps to improve the \emph{local} microstructure in the translated image while 
larger subsamples rather keep \emph{global} information.
Finally, we want to note that we have verified the generalization of our 
procedure by training the NN only on a smaller part of the high-resolution 
image, but finally transforming the whole one.
By this procedure, there were areas in the big image which the NN has never 
seen during training.
In all our test cases, the procedure generalized very well with no visible 
artifacts or mis-translations as long as no crucial image content had been cut 
off for the training process.

\section{Conclusion}
\label{sec:conclusion}

In this paper, we have introduced a method which allows to perform unpaired 
domain translation on high-resolution images.
It is based on the idea not to process the whole image at once but to 
apply the training and evaluation on subsamples of random size (and/or 
position) which were downscaled to a fixed small image size.
With this method, we were able to create high-quality domain translations which 
fulfill micro- and macro-consistency, and preserve image details well.
We performed training and evaluation of the GAN setting on a ``small'' standard 
desktop GPU which underlines that high-resolution domain transfer does not 
require large and expensive GPU hardware or clusters.
We have applied the method to various domain translations covering the 
range from very similar to very different domains.

We see potential applications of this procedure especially in the case of 
high-quality and high-resolution (test) data generation for different use 
cases as, \eg, autonomous driving.

We generally propose to apply the idea of processing data with NNs not 
as a whole but on (random) overlapping subsamples also for different kinds of 
data types and in other application fields.
For example, one might use similar approaches in the field of 3-dimensional 
objects or meshes 
\cite{3dgan,jiang2017,chang2015} 
where parts of the object could be processed separately.
Another field can be graph data 
\cite{dai2016,dai2017,hamilton2017,grover2016} 
where one could work on single subgraphs instead of the whole big one.

\section*{Appendix}

\subsection*{Neural Network Architecture}

For the domain translations performed in this paper, a \textsc{Unit}-like GAN 
architecture \cite{UNIT} with shared latent space has been used with the 
generators and discriminators set up as listed in Tab.~\ref{tab:net-config}.

\begin{table}[h!]
\centering \small
\begin{tabular}{lrrr} 
\toprule
\textbf{Generator} & Filter size (num), Norm, Activ.\\
\midrule
Down-Convolution & $7\times7$  (64), --, LeakyReLU \\
Down-Convolution & $3\times3$ (128), --, LeakyReLU \\
Down-Convolution & $3\times3$ (256), --, LeakyReLU \\
Down-Convolution & $3\times3$ (512), --, LeakyReLU \\
Residual ($3\times$) & $3\times3$ (512), Inst.-Norm, ReLU \\
Residual ($2\times$, shared) & $3\times3$ (512), Inst.-Norm, ReLU \\
Residual ($3\times$) & $3\times3$ (512), Inst.-Norm, ReLU \\
Up-Convolution & $3\times3$ (256), --, LeakyReLU \\
Up-Convolution & $3\times3$ (128), --, LeakyReLU \\
Up-Convolution & $3\times3$  (64), --, LeakyReLU \\
Up-Convolution & $3\times3$   (3), --, Tanh\\
\bottomrule \\
\toprule
\textbf{Discriminator} & Filter size (num), Norm, Activ.\\
\midrule
Down-Convolution & $3\times3$  (64), --, LeakyReLU \\
Down-Convolution & $3\times3$ (128), --, LeakyReLU \\
Down-Convolution & $3\times3$ (256), --, LeakyReLU \\
Down-Convolution & $3\times3$ (512), --, LeakyReLU \\
Down-Convolution & $1\times1$   (1), --, LeakyReLU \\
\bottomrule
\end{tabular}
\caption{%
Network architecture of the GAN's generator and discriminator used for the 
images presented in this paper.
}
\label{tab:net-config}
\end{table}

\subsection*{Image details}

\begin{figure*}[p]
Image details of \FIG~\ref{fig:landscape2}\\
\includegraphics[width=.12\textwidth]{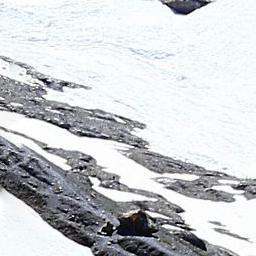}
\includegraphics[width=.12\textwidth]{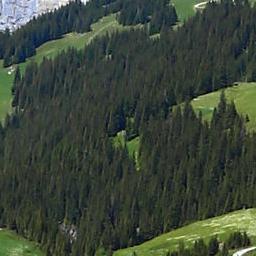}
\includegraphics[width=.12\textwidth]{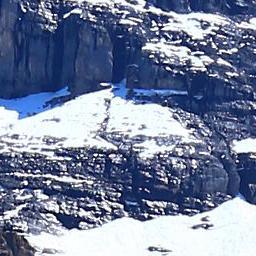}
\includegraphics[width=.12\textwidth]{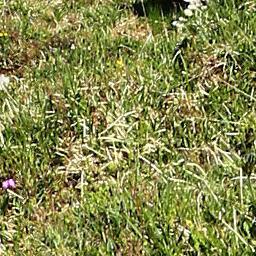} 
\includegraphics[width=.12\textwidth]{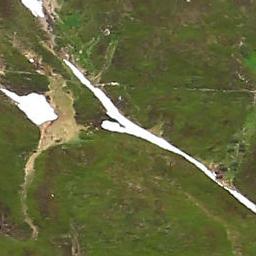}
\includegraphics[width=.12\textwidth]{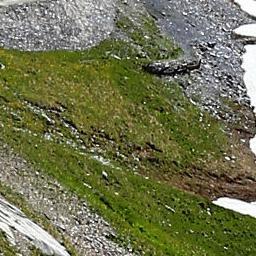}
\includegraphics[width=.12\textwidth]{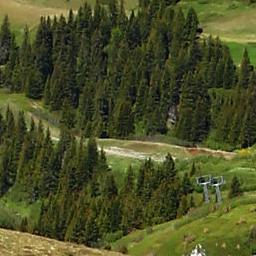}
\includegraphics[width=.12\textwidth]{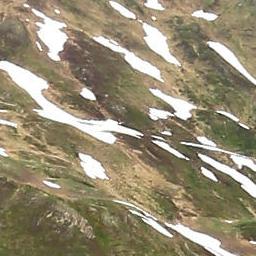} 
\\
\includegraphics[width=.12\textwidth]{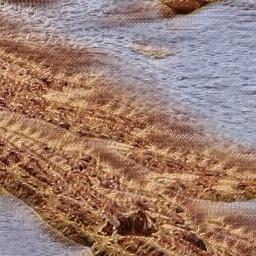}
\includegraphics[width=.12\textwidth]{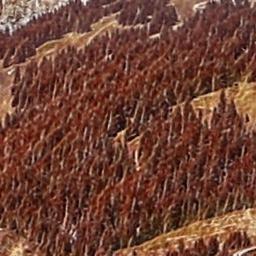}
\includegraphics[width=.12\textwidth]{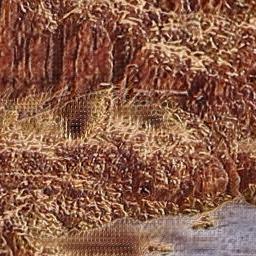}
\includegraphics[width=.12\textwidth]{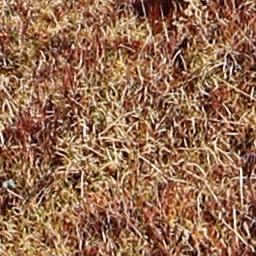} 
\includegraphics[width=.12\textwidth]{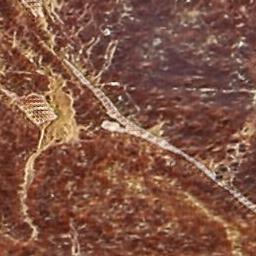}
\includegraphics[width=.12\textwidth]{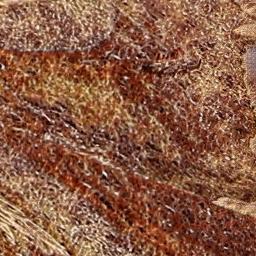}
\includegraphics[width=.12\textwidth]{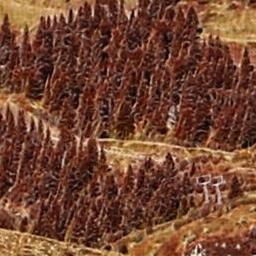}
\includegraphics[width=.12\textwidth]{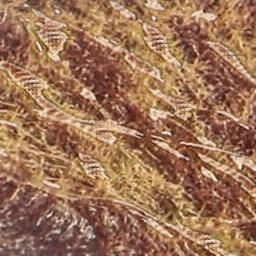} 
\\[2em]
Image details of \FIG~\ref{fig:street1}\\
\includegraphics[width=.12\textwidth]{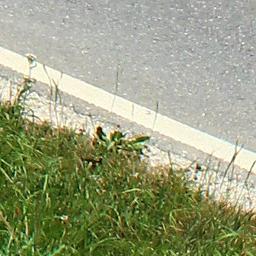}
\includegraphics[width=.12\textwidth]{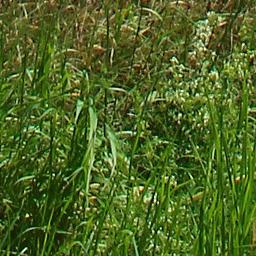}
\includegraphics[width=.12\textwidth]{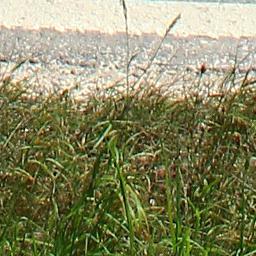}
\includegraphics[width=.12\textwidth]{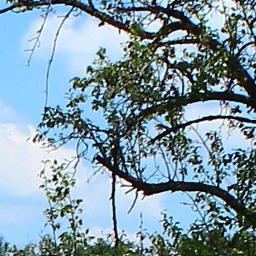} 
\includegraphics[width=.12\textwidth]{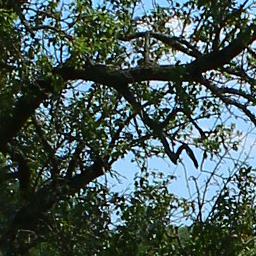}
\includegraphics[width=.12\textwidth]{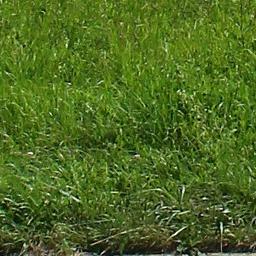}
\includegraphics[width=.12\textwidth]{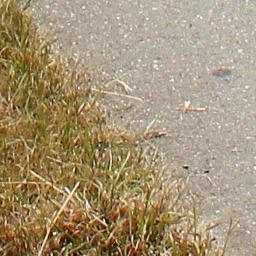}
\includegraphics[width=.12\textwidth]{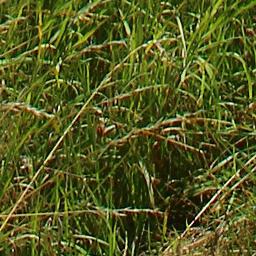}  
\\
\includegraphics[width=.12\textwidth]{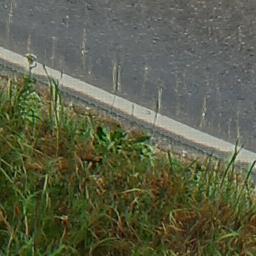}
\includegraphics[width=.12\textwidth]{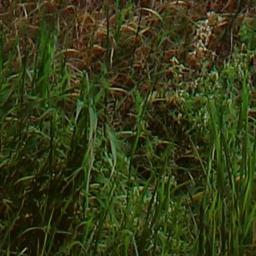}
\includegraphics[width=.12\textwidth]{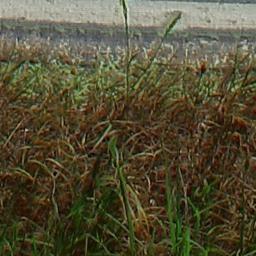}
\includegraphics[width=.12\textwidth]{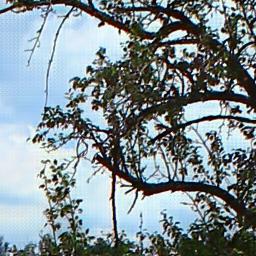} 
\includegraphics[width=.12\textwidth]{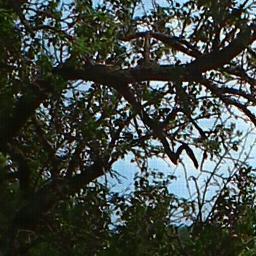}
\includegraphics[width=.12\textwidth]{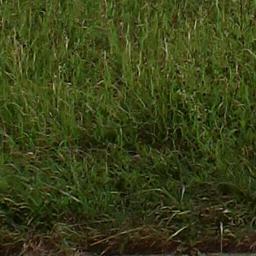}
\includegraphics[width=.12\textwidth]{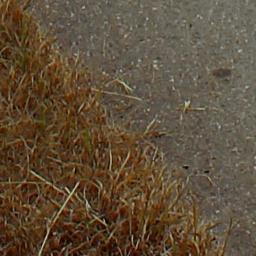}
\includegraphics[width=.12\textwidth]{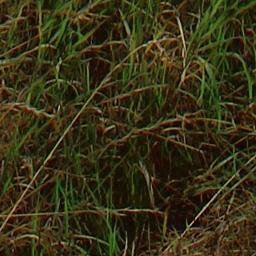} 
\\[2em]
Image details of \FIG~\ref{fig:street2}\\
\includegraphics[width=.12\textwidth]{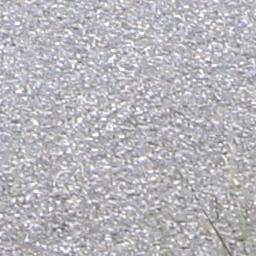}
\includegraphics[width=.12\textwidth]{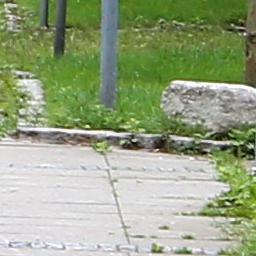}
\includegraphics[width=.12\textwidth]{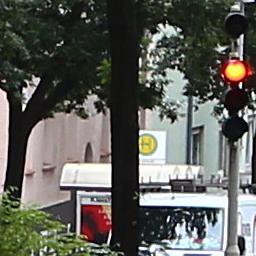}
\includegraphics[width=.12\textwidth]{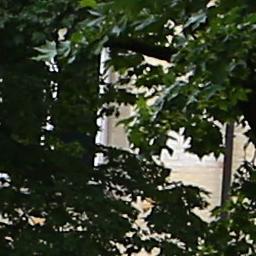} 
\includegraphics[width=.12\textwidth]{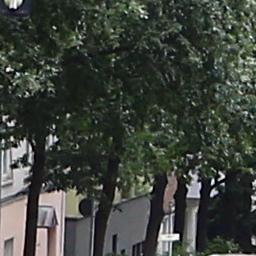}
\includegraphics[width=.12\textwidth]{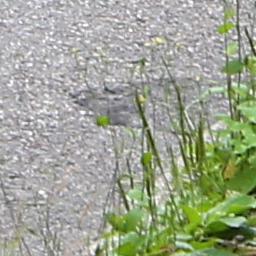}
\includegraphics[width=.12\textwidth]{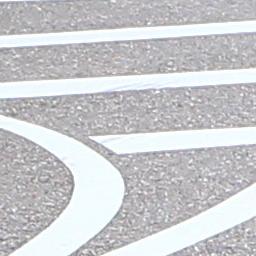}
\includegraphics[width=.12\textwidth]{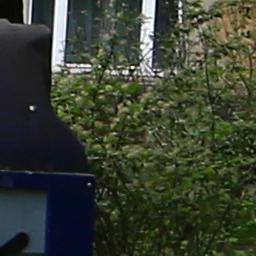} 
\\
\includegraphics[width=.12\textwidth]{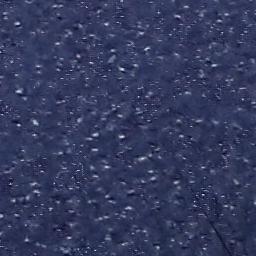}
\includegraphics[width=.12\textwidth]{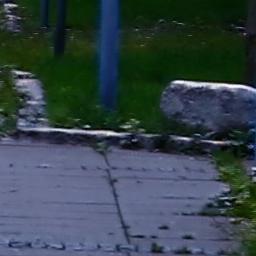}
\includegraphics[width=.12\textwidth]{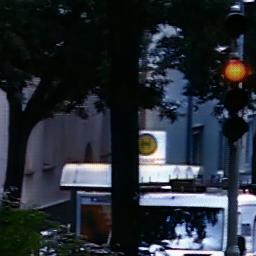}
\includegraphics[width=.12\textwidth]{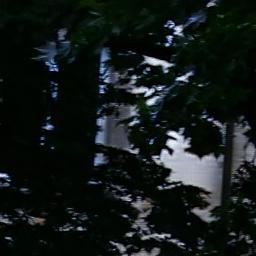} 
\includegraphics[width=.12\textwidth]{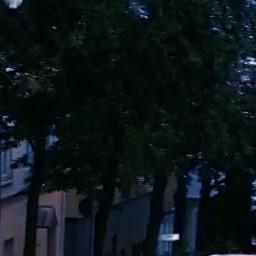}
\includegraphics[width=.12\textwidth]{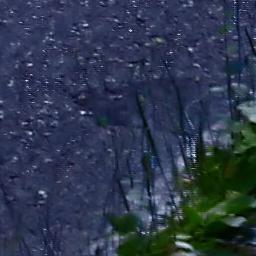}
\includegraphics[width=.12\textwidth]{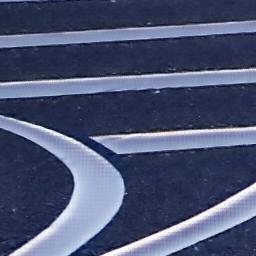}
\includegraphics[width=.12\textwidth]{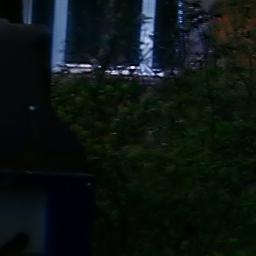} 
\\[2em]
Image details of \FIG~\ref{fig:street5}\\
\includegraphics[width=.12\textwidth]{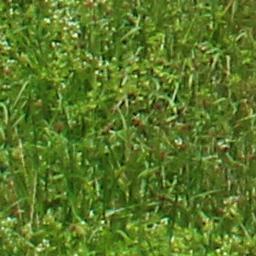}
\includegraphics[width=.12\textwidth]{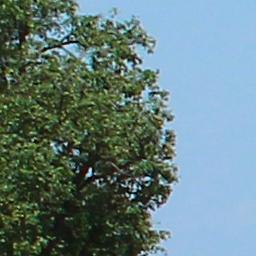}
\includegraphics[width=.12\textwidth]{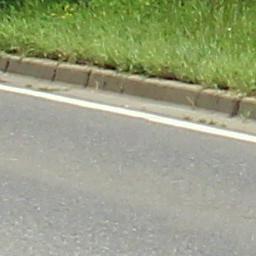}
\includegraphics[width=.12\textwidth]{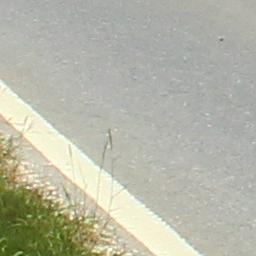} 
\includegraphics[width=.12\textwidth]{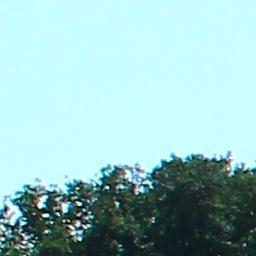}
\includegraphics[width=.12\textwidth]{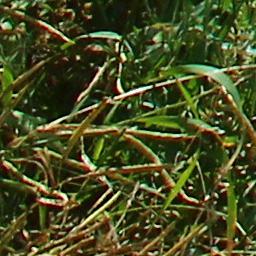}
\includegraphics[width=.12\textwidth]{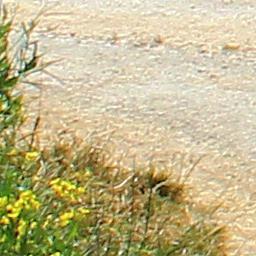}
\includegraphics[width=.12\textwidth]{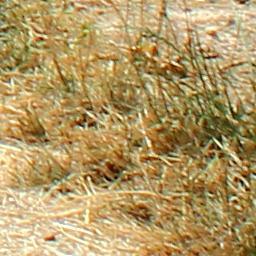} 
\\
\includegraphics[width=.12\textwidth]{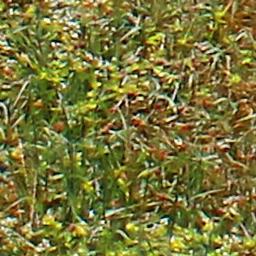}
\includegraphics[width=.12\textwidth]{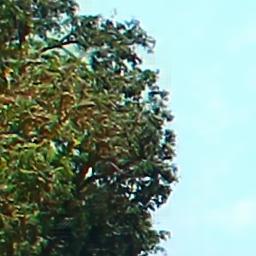}
\includegraphics[width=.12\textwidth]{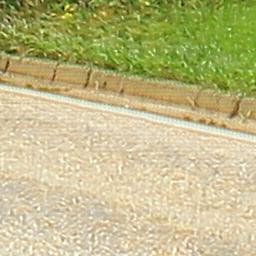}
\includegraphics[width=.12\textwidth]{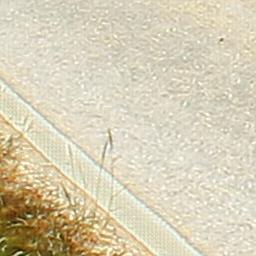} 
\includegraphics[width=.12\textwidth]{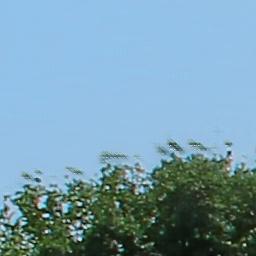}
\includegraphics[width=.12\textwidth]{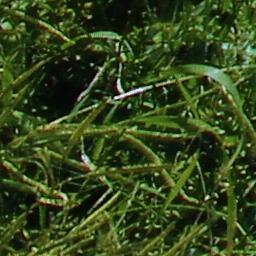}
\includegraphics[width=.12\textwidth]{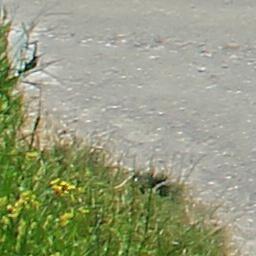}
\includegraphics[width=.12\textwidth]{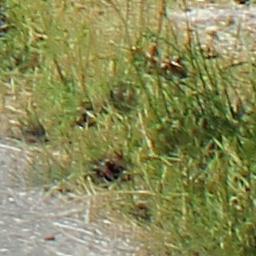} 
\caption{%
Details of the original and transformed images in \FIGS~\ref{fig:landscape2}, 
\ref{fig:street1}, \ref{fig:street2}, and \ref{fig:street5} presented in their 
full resolution.
}
\label{fig:landscape2_details}
\end{figure*}

Table~\ref{tab:resolutions} summarizes the image geometries and resolutions of 
the images presented in this paper.
Further image details are presented in \FIG~\ref{fig:landscape2_details}.

%%%%%%%%%%%%%%%%%%%%%%%%%%%%%%%%%%%%%%%%%%%%%%%%%%%%%%%%%%%%%%%%%%%%%%%%%%%%%%%%
\begin{table}[h!]
\centering \small
\begin{tabular}{rrrr}
\toprule
Figure & Width [px] & Height [px] & Total [Mio. px] \\
\midrule
3   &   15,884 & 3,271 & 51.96\\ % landscape2
4   &   12,895 & 3,472 & 44.77\\ % street2
5   &    5,472 & 3,648 & 19.96\\ % street3
6   &    5,472 & 3,648 & 19.96\\ % street3
7   &    2,448 & 2,448 &  5.99\\ % trafficsigns
\bottomrule
\end{tabular}
\caption{%
Resolution of the original and transformed images in 
\FIGS~\ref{fig:landscape2}--\ref{fig:trafficsign} and the 
corresponding total number of pixels.
}
\label{tab:resolutions}
\end{table}
%%%%%%%%%%%%%%%%%%%%%%%%%%%%%%%%%%%%%%%%%%%%%%%%%%%%%%%%%%%%%%%%%%%%%%%%%%%%%%%%

%%%%%%%%%%%%%%%%%%%%%%%%%%%%%%%%%%%%%%%%%%%%%%%%%%%%%%%%%%%%%%%%%%%%%%%%%%%%%%%%
%%%%%%%%%%%%%%%%%%%%%%%%%%%%%%%%%%%%%%%%%%%%%%%%%%%%%%%%%%%%%%%%%%%%%%%%%%%%%%%%

\footnotesize
% \bibliography{GANs,styleTransfer,graph,anomalydetection,objectdetection,nlp,%
% autoencoder}
% \bibliographystyle{unsrt}

\end{document}